\documentclass[format=acmsmall, review=false, screen=true]{acmart}
\usepackage[utf8]{inputenc} 
\usepackage[T1]{fontenc}    
\usepackage{hyperref}       
\usepackage{url}            
\usepackage{booktabs}       
\usepackage{amsfonts}       
\usepackage{nicefrac}       
\usepackage{microtype}      
\usepackage{xcolor}         
\usepackage{multirow}
\usepackage{graphicx}
\usepackage{lscape}
\usepackage{booktabs}
\usepackage{wrapfig,lipsum,booktabs}
\usepackage{makecell}
\usepackage{amsmath}
\usepackage{graphicx}
\usepackage{subcaption}
\usepackage{enumitem}
\newtheorem{proposition}{Proposition}
\AtBeginDocument{%
  }

\setcopyright{acmlicensed}
\copyrightyear{2018}
\acmYear{2018}
\acmDOI{XXXXXXX.XXXXXXX}
\acmJournal{TKDD}

\acmConference[Conference acronym 'XX]{Make sure to enter the correct
  conference title from your rights confirmation emai}{June 03--05,
  2018}{Woodstock, NY}

\acmISBN{978-1-4503-XXXX-X/18/06}

\begin{document}

\def\method{DuSEGO}
\title{\method{}: Dual Second-order Equivariant Graph Ordinary Differential Equation}

\author{Yingxu Wang}
\affiliation{
  \institution{Mohamed bin Zayed University of Artificial Intelligence}
  \city{Abu Dhabi}
  \country{United Arab Emirates}
}
\email{yingxv.wang@gmail.com}

\author{Nan Yin}
\affiliation{
\institution{Mohamed bin Zayed University of Artificial Intelligence}
\city{Abu Dhabi}
  \country{United Arab Emirates}
}
\email{yinnan8911@gmail.com}

\author{Mingyan Xiao}
\affiliation{
\institution{California State Polytechnic University}
\city{Pomona}
  \country{United States}
}
\email{mxiao@cpp.edu}

\author{Xinhao Yi}
\affiliation{
\institution{University of Glasgow}
  \city{Glasgow}
  \country{Scotland}
}
\email{x.yi.2@research.gla.ac.uk}

\author{Siwei Liu\textsuperscript{\dag}}
\affiliation{
\institution{Mohamed bin Zayed University of Artificial Intelligence}
\city{Abu Dhabi}
  \country{United Arab Emirates}
}
\email{Siwei.Liu@mbzuai.ac.ae}

\author{Shangsong Liang\textsuperscript{\dag}}
\affiliation{
\institution{Sun Yat-sen University, and Mohamed bin Zayed University of Artificial Intelligence}
  \country{China and United Arab Emirates}
  \city{Guangzhou and Abu Dhabi}
}
\email{liangshangsong@gmail.com}




\thanks{\textsuperscript{\dag}Corresponding authors.}

\begin{abstract}
Graph Neural Networks (GNNs) with equivariant properties have achieved significant success in modeling complex dynamic systems and molecular properties. However, their expressiveness ability is limited by: (1) Existing methods often overlook the over-smoothing issue caused by traditional GNN models, as well as the gradient explosion or vanishing problems in deep GNNs. (2) Most models operate on first-order information, neglecting that the real world often consists of second-order systems, which further limits the model's representation capabilities. To address these issues, we propose the \textbf{Du}al \textbf{S}econd-order \textbf{E}quivariant \textbf{G}raph \textbf{O}rdinary Differential Equation (\method{}) for equivariant representation. Specifically, \method{} applies the dual second-order equivariant graph ordinary differential equations (Graph ODEs) to both graph embeddings and node coordinates simultaneously. Theoretically, we first prove that \method{} maintains the equivariant property. Furthermore, we provide theoretical insights showing that \method{} effectively alleviates the over-smoothing problem in both feature representation and coordinate update. Additionally, we demonstrate that the proposed \method{} mitigates the exploding and vanishing gradients problem, facilitating the training of deep multi-layer GNNs. Extensive experiments on benchmark datasets validate the superiority of the proposed \method{} compared to baselines. 
\end{abstract}



\received{20 February 2007}
\received[revised]{12 March 2009}
\received[accepted]{5 June 2009}

\maketitle
\section{Introduction}

GNNs~\cite{kipf2017semisupervised} with equivariant properties~\cite{satorras2021n,han2022equivariant,yin2023coco,pang2023sa} have become essential tools in modeling complex dynamic systems~\cite{meng2024deep,huang2021equivariant,liu2024dynamic} and molecular properties~\cite{brandstetter2021geometric} due to their ability to capture symmetrical relationships within data. Equivariant GNNs are particularly effective in scenarios when maintaining consistency under transformations (e.g., rotations, reflections) is crucial. The consistency of equivariant representations allows these models to accurately reflect the inherent properties of the systems they are designed to study, leading to more reliable and insightful predictions, which is now an essential topic in machine learning~\cite{battaglia2016interaction,yin2022deal,kohler2019equivariant}.

\begin{figure}
    \centering
    \begin{subfigure}[b]{0.48\textwidth} 
        \centering
        \includegraphics[width=\textwidth]{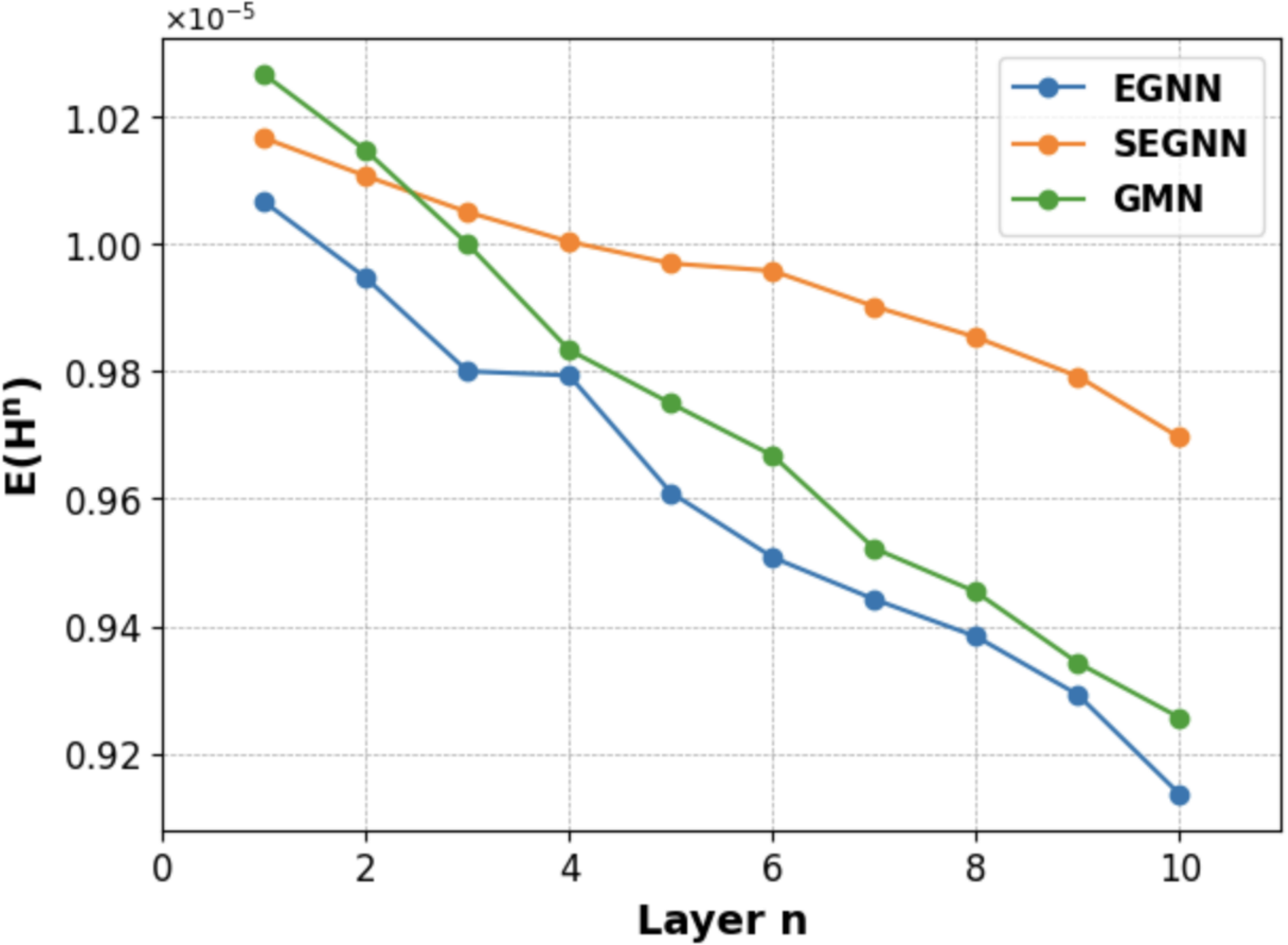}
        \caption{Dirichlet energy $\mathbf{E}\left(\mathbf{H}^n\right)$ of node features.}
        \label{fig:sub1}
    \end{subfigure}%
    \hfill 
    \begin{subfigure}[b]{0.48\textwidth}
        \centering
        \includegraphics[width=0.98\textwidth,height=0.738\textwidth]{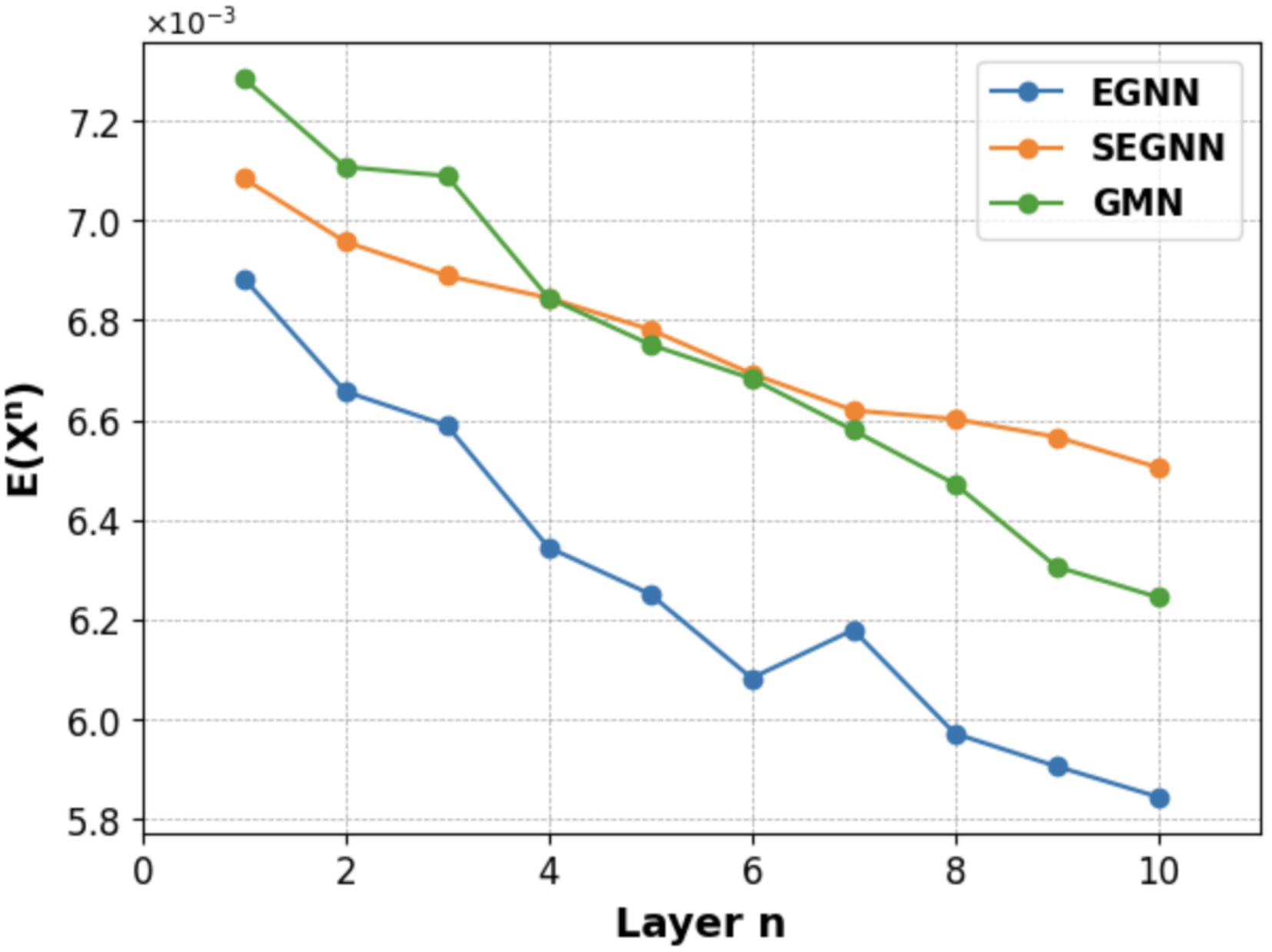}
        \caption{Dirichlet energy $\mathbf{E}\left(\mathbf{X}^n\right)$ of node coordinates.}
        \label{fig:sub2}
    \end{subfigure}
    \caption{Dirichlet energy $\mathbf{E}\left(\mathbf{H}^n\right)$ of node features $\mathbf{H}^n$ and $\mathbf{E}\left(\mathbf{X}^n\right)$ of node coordinates $\mathbf{X}^n$ propagated through a EGNN \cite{satorras2021n}, SEGNN \cite{brandstetter2021geometric} and GMN \cite{huang2021equivariant} on a N-body system dataset, where we give the definition of the Dirichlet energy in Eq.~\ref{energy}.}
    \label{fig:oversmoothing}
    \vspace{-0.5cm}
\end{figure}

Despite their successes, existing Equivariant GNNs face notable limitations that constrain their performance and expressiveness. First, \textit{the over-smoothing problem}~\cite{li2018deeper,keriven2022not,min2020scattering,zhang2023investigating}. The classical Equivariant GNNs (such as EGNN~\cite{satorras2021n} and SEGNN~\cite{brandstetter2021geometric}) typically utilize the GNNs~\cite{kipf2017semisupervised,yin2022dynamic} as a backbone, as well as cooperating coordinates for equivariant representation. However, GNNs usually suffer from the over-smoothing problem, leading to the indistinguishable representation of features and coordinates, which is shown in Fig. \ref{fig:oversmoothing}. Second, \textit{gradient explosion or vanishing problems}~\cite{dasoulas2021lipschitz,guo2022orthogonal,yin2022generic,zhang2022would}. Motivated by the notion that larger models are usually more expressive~\cite{kaplan2020scaling,ai2023gcn}, we tend to explore the expressiveness ability by enlarging the model scale. However, as the network depth increases, gradients can either become excessively large or diminish to near zero, which limits the scalability and depth of GNNs.
Third, \textit{real-world simulation}~\cite{norcliffe2020second}. 
Most Equivariant GNNs~\cite{shou2023adversarial,thomas2018tensor,tang2024merging,yinsport} primarily focus on first-order information. However, many physical systems are governed by second-order laws~\cite{norcliffe2020second}. The first-order methods overlook the higher-order interactions that are critical in real-world systems, and limit the representational ability. Even though some works~\cite{huang2021equivariant,liu2024segno} try to introduce second-order information into equivariant learning, they simply focus on simulating the position in a dynamic physical system, ignoring the expressive representation for molecular properties prediction.

To address these challenges, we take an insight into the second-order method in Equivariant GNNs and a framework named \textbf{Du}al \textbf{S}econd-order \textbf{E}quivariant \textbf{G}raph \textbf{O}rdinary Differential Equation (\method{}). The proposed \method{} is a universal framework that transfers the first-order Equivariant GNNs to second-order, which aims to enhance Equivariant GNNs' expressiveness and address the issues of over-smoothing and gradient stability. First, we prove that the proposed \method{} maintains the equivariant properties. Then, we analyze the properties of \method{} to reveal how the proposed method solves the over-smoothing, gradient exploding, and vanishing problems, facilitating the training of deep GNNs.
Extensive experiments on benchmark datasets validate the superiority of the proposed \method{} compared to existing baselines.

The main contributions can be summarized as threefold:
\begin{itemize}[itemsep=2pt,topsep=0pt,parsep=0pt]
\item The proposed \method{} is flexible to accommodate the first-order equivariant graph ODE methods to the second-order, which still maintains the equivariant properties.
\item We analyze the properties of \method{}, and demonstrate that the second-order graph ODE can effectively mitigate the over-smoothing, gradient exploding, and vanishing problems.
\item Extensive experiments conducted on various real-world datasets demonstrate the effectiveness of the proposed \method{}.
\end{itemize}
\section{Related Work}

\textbf{Equivariant Graph Neural Networks.} Early graph learning methods explored spectral and hypergraph-based techniques that emphasize global topology and higher-order relationships, particularly for tasks such as semi-supervised classification and structure-aware clustering~\cite{yin2024continuous,ju2024survey,wang2025nested}. Building upon these insights, GNNs~\cite{wang2025dynamically,yin2025dream} with equivariant properties have demonstrated their effectiveness in various tasks, especially for 3D molecular graphs~\cite{xian2025molrag} and physical symmetry of systems \cite{zhang2023universal}. The Tensor Field Networks (TFNs)~\cite{thomas2018tensor} method was first proposed for 3D point clouds, which only leverage the positional information of nodes without any additional features. SE(3)-transformers \cite{fuchs2020se} deploy an attention mechanism into TFNs to improve the model performance. EGNN \cite{satorras2021n}, a simple but effective form of equivariant GNNs, was proposed and has been the backbone of the following research. SEGNN \cite{brandstetter2021geometric} incorporates geometric and physical properties into the message-passing framework, significantly enhancing the accuracy and predictive capabilities of EGNN in understanding three-dimensional molecular structures. 
However, these methods focus on first-order velocity information, even though second-order motion laws govern many physical systems.
GMN \cite{huang2021equivariant} first introduced the second-order information into equivariant learning, and SEGNO \cite{liu2024segno} further applied the second-order ODE for solving the physical dynamic system. However, these methods only target the description of positional state information in the physical world and cannot be applied to learning molecular feature representations. The proposed \method{} utilizes the second-order ODE to evaluate the evolution of equivariant representation and position, making it suitable for applications in dynamic physical systems and molecular properties prediction.

\textbf{Graph Ordinary Differential Equations.} 
Graph Ordinary Differential Equations (Graph ODEs) \cite{poli2019graph,ijcai2025p396,yin2023messages} draw inspiration from Neural Ordinary Differential Equations \cite {chen2018neural} and provide a sophisticated mechanism for modeling dynamic changes within graph structures, showing great potential in graph learning \cite{yin2023dream,ijcai2025p219,sanchez2019hamiltonian,yin2023omg}. A series of works proposes to deploy first-order or second-order ODEs into GNNs to enhance their expressiveness. LG-ODE \cite{huang2020learning} learns the dynamics of continuous systems by using models that can handle irregular sampling and incomplete data, improving prediction accuracy and robustness. CG-ODE \cite{huang2021coupled} uses coupled ordinary differential equations on graphs to learn and model the dynamics of interacting systems effectively. HOPE \cite{luo2023hope} leverages second-order differential equations to capture complex interactions and dynamic behaviors within graph structures. GraphCon \cite{rusch2022graph} proposes a novel framework that integrates the dynamics of coupled oscillators with graph theory, analyzing stability across complex networked systems. However, the above works do not consider the equivariance inherent in the graph structure, which is challenging to extend to 3D molecular graphs and the physical symmetries of systems.
\section{Preliminary}

\textbf{Equivariance} 
Let $T_g : X\rightarrow X$ be a set of transformations on $X$ for the abstract group $g \in G$. We say a function $\phi : X \rightarrow Y$ is equivariant to $g$ if there exists an equivalent transformation on its output space $S_g : Y \rightarrow Y$ such that: $\phi(T_g(\mathbf{x}))=S_g(\phi(\mathbf{x}))$. The equivariance properties include translations, rotations, and permutations, which ensure that the model's predictions are consistent regardless of the orientation or position of objects within the input data.  
\begin{itemize}[itemsep=2pt,topsep=0pt,parsep=0pt]
    \item \textit{Translation:} Let $\mathbf{x}+g$ be shorthand for $(\mathbf{x_1}+g,\cdots,\mathbf{x_M}+g)$, then $\phi(x)+g=\phi(x+g)$.
    \item  \textit{Rotation:} Let $Q\mathbf{x}$ be shorthand for $(Q\mathbf{x_1},\cdots,Q\mathbf{x_M})$, then $Q\phi(\mathbf{x})=\phi(Q\mathbf{x})$.
\end{itemize}

\textbf{Graph Nerual Networks}
Given graph $G=(\mathcal{V}, \mathcal{E})$, let $\mathbf{h}_v^{(k)}$ denote embedding vector for node $v$ at the $l$-th layer. For node $v \in \mathcal{V}$, we aggregate the embeddings from the neighbors of $v$ at layer $l-1$. Then, the node embedding $\mathbf{h}_v^{(k)}$ is iteratively updated by combining the embedding of $v$ in the previous layer with the aggregated neighbor embedding. In formulation, 
$\mathbf{h}_v^{l}$ is calculated as follows:
\begin{equation}
\mathbf{h}_{v}^{l}= \mathcal{C}^{(k)}\left(\mathbf{h}_{v}^{l-1}, \mathcal{A}^{(k)} \left(\left\{\mathbf{h}_{u}^{l-1}\right\}_{u \in \mathcal{N}(v)}\right) \right),
\end{equation}
where $\mathcal{N}(v)$ represents the neighbors of $v$. $\mathcal{A}^{l}$ and $\mathcal{C}^{l}$ represent the aggregation and combination operations at the $l$-th layer, respectively. At last, we summarize all node representations at the $L$-th layer with a readout function into the graph-level representation, which can be formulated as follows:
\begin{equation}
\mathbf{z}= F\left(G\right)=\operatorname{READOUT}\left(\left\{\mathbf{h}_{v}^{L}\right\}_{v \in \mathcal{V}}\right),
\end{equation}
where $\mathbf{z}$ is the graph-level representation of $G$ and $\mathbf{\theta_e}$ denotes the parameter of our GNN-based encoder. $L$ denotes the number of the graph convolutional layers. The readout function can be implemented in different ways, such as summarizing all nodes' representations~\cite{xu2018powerful,wang2025nested} or using a virtual node~\cite{li2015gated}.

\textbf{Equivariant GNNs} Consider a graph $\mathcal{G}=(\mathcal{V}, \mathcal{E})$ with nodes $v_i \in \mathcal{V}$ and edges $e_{i j} \in \mathcal{E}$. In addition to the feature node embeddings $\mathbf{h}_i \in \mathbb{R}^{d}$, the $n$-dimensional coordinate $\mathbf{x}_i \in \mathbb{R}^{n}$ associated with each of the graph nodes are also considered, where $d$ is the dimension of node feature, and $n$ is the dimension of node position. A GNN with Equivariance $\phi_{\theta}$ with parameter $\theta$ can complete given tasks by message passing layers. Specially, each layer of $\phi_{\theta}$ can be formulated as follows:
\begin{equation}
\label{equivariant}
\boldsymbol{m}_{i j}=\mu\left(\mathbf{h}_i^l, \mathbf{h}_j^l, \mathbf{x}_i^l, \mathbf{x}_j^l, e_{i j}\right), \quad \mathbf{x}_i^{l+1}, \mathbf{h}_i^{l+1} =\nu\left(\mathbf{h}_i^l, \mathbf{x}_i^l, \sum_{j \in \mathcal{N}(i)} \mathbf{m}_{i j}\right),
\end{equation}
where $\mu$ and $\nu$ are the edge message function and node update function, $\mathcal{N}(i)$ are the neighbors of node $i$, $\mathbf{x}_i^{l} \in \mathbb{R}^{n} $ and $\mathbf{h}_i^{l} \in \mathbb{R}^{d} $ are the updated feature and position of node $i$. For containing E(n) equivariance for each layer, $\mu$ and $\nu$ could be both equivariant (e.g., SEGNN \cite{brandstetter2021geometric}) or alternatively, $\mu$ is equivariant and $\nu$ is invariant (e.g., EGNN \cite{satorras2021n}).

\textbf{Dirichlet Energy} As introduced in~\cite{rusch2022graph},
let $\mathbf{X}^n$ denote embeddings of node on the $n$-th layer. The Dirichlet energy function is defined as:
\begin{equation}
    E(\mathbf{X}^n)\leq C_1 \exp^{-C_2n},
    \label{energy}
\end{equation}
where $C_1, C_2>0$ are some constant numbers. 

\section{Method}

In this section, we introduce the proposed \method{}. In the dynamic systems and molecular properties prediction, we utilize the second-order ODE to simulate the representation and trajectory.

\subsection{Second-order Graph ODE}
To model high-order correlations in long-term temporal trends, ~\cite{rusch2022graph} first proposed the second-order graph ODE, which is represented as:
\begin{equation}
\label{second}
    \mathbf{X}^{''}=\sigma(\mathbf{F}_{\theta}(\mathbf{X},t))-\gamma \mathbf{X}-\alpha \mathbf{X}^{'},
\end{equation}
where $\left(\mathbf{F}_\theta(\mathbf{X},t)\right)_i=\mathbf{F}_\theta \left(\mathbf{X}_i(t),\mathbf{X}_j(t),t\right)$ is a learnable coupling function with parameters $\theta$. $\sigma$ is the mapping function, which is implemented with multiplayer perceptions. Due to the unavailability of an analytical solution for Eq.~\ref{second}, GraphCON~\cite{rusch2022graph} addresses it through an iterative numerical solver employing a suitable time discretization method. GraphCON utilizes the IMEX (implicit-explicit) time-stepping scheme, {an extension of the symplectic Euler method~\cite{wanner1996solving} that accommodates systems with an additional damping term.} 
\begin{equation}
\label{second_graph_ode}
\begin{aligned}
    \mathbf{Y}^n=\mathbf{Y}^{n-1}+\Delta t&\left[\sigma(\mathbf{F}_\theta(\mathbf{X}^{n-1},t^{n-1}))
    -\gamma\mathbf{X}^{n-1}-\alpha\mathbf{Y}^{n-1}\right],\\ \mathbf{X}^n=&\mathbf{X}^{n-1}+\Delta t\mathbf{Y}^n,\, n=1,\cdots,N,
\end{aligned}
\end{equation}
where $\Delta t>0$ is a fixed time-step and $\mathbf{Y}^n$, $\mathbf{X}^n$ denote the hidden node features at time $t^n=n\Delta t$.

\subsection{\method{}}

To efficiently evaluate the graph representation and the coordinate information, we present the framework in a graph dynamical system by following the second-order ODEs:
\begin{equation}
\label{eq:second_ode}
    \mathbf{X}''=\sigma (\mathbf{F}_\theta(\mathbf{X},\mathbf{H},t))-\gamma_1 \mathbf{X}-\alpha \mathbf{X}',\quad
    \mathbf{H}''=\sigma (\mathbf{F}_\theta(\mathbf{X},\mathbf{H},t))-\gamma_2 \mathbf{H}-\alpha \mathbf{H}',
\end{equation}
where $\mathbf{X}$ is the coordinate embeddings and $\mathbf{H}$ is the node features.
Similar with Eq.~\ref{second_graph_ode}, we have:
\begin{equation}
\label{ode1}
    \mathbf{Y}^n=\mathbf{Y}^{n-1}+\Delta t\left[\sigma(\mathbf{F}_\theta(\mathbf{X}^{n-1},\mathbf{H}^{n-1},t^{n-1}))
    -\gamma_1\mathbf{X}^{n-1}-\alpha\mathbf{Y}^{n-1}\right], \mathbf{X}^n=\mathbf{X}^{n-1}+\Delta t\mathbf{Y}^n,\, n=1,\cdots,N,
\end{equation}
and 
\begin{equation}
\label{ode2}
    \mathbf{U}^n=\mathbf{U}^{n-1}+\Delta t\left[\sigma(\mathbf{F}_\theta(\mathbf{X}^{n-1},\mathbf{H}^{n-1},t^{n-1}))
    -\gamma_2\mathbf{H}^{n-1}-\alpha\mathbf{U}^{n-1}\right], \mathbf{H}^n=\mathbf{H}^{n-1}+\Delta t\mathbf{U}^n,\, n=1,\cdots,N,
\end{equation}

To keep the Equivariant properties of \method{}, the proposed method allows for any Equivariant GNNs to be used as $\mathbf{F}_\theta$. 

\textbf{Choice of function $\mathbf{F}_\theta$}.
Our framework allows for any learnable equivariant GNNs to be used as $\mathbf{F}_\theta$. 
In this paper, we focus on the two particularly popular choices:

$\textit{Choice 1: Equivariant Graph Neural Networks (EGNNs)}$:
\begin{equation}
    \mathbf{F}_\theta(\mathbf{X}^l,\mathbf{H}^l,t^l)=\textbf{h}_i^{l+1},\;\mathbf{x}_i^{l+1}=\phi_h(\textbf{h}_i^l,\textbf{m}_i),\; \mathbf{x}_i^l+C\sum_{j\neq i}(\mathbf{x}_i^l-\mathbf{x}_j^l)\phi_x(\mathbf{m}_{ij}),
\end{equation}
where 
$\mathbf{m}_{ij}=\phi_e(\mathbf{h}_i^l,\mathbf{h}_j^l,||\mathbf{x}_i^l-\mathbf{x}_j^l||^2,a_{ij})$ and
$\textbf{m}_i=\sum_{j\neq i}\textbf{m}_{ij}$. $\phi_e$ and $\phi_h$ are the edge and node operations, respectively, which are commonly approximated by Multilayer Perceptrons (MLPs). Nodes $j$ are the neighbors of node $i$.

\textit{Choice 2: Steerable E(3) Equivariant Graph Neural Networks (SEGNN)}:

\begin{equation}
\label{eq:update}
\mathbf{F}_\theta(\mathbf{X}^l,\mathbf{H}^l,t^l)=\phi_h(\mathbf{h}_i^l,\sum_{j\in\mathcal{N}(i)}\mathbf{m}_{ij},\mathbf{a}_i),\; \mathbf{m}_{ij}=\phi_m(\mathbf{h}_i,\mathbf{h}_j,||\mathbf{x}_j-\mathbf{x}_i||^2,\mathbf{a}_{ij}),
\end{equation}
where $||\mathbf{x}_j-\mathbf{x}_i||^2$ is the squared relative distance between two nodes $v_i$ and $v_j$, $\phi_m$ and $\phi_h$ are O(3) steerable MLPs, $\mathbf{a}_{ij}$ and $\mathbf{a}_i$ are steerable edge and node attributes, $\mathbf{a}_i=\frac{1}{|\mathcal{N}(I)|}\sum_{j\in\mathcal{N}(i)}\mathbf{a}_{ij}$. $\mathcal{N}(i)$ denotes the neighbors of node $v_i$.

\textbf{Steady States of \method{}}.
It is straightforward to see that the steady states $\mathbf{X}^*$,$\mathbf{H}^*$ of the \method{} dynamical system with an autonomous function $\mathbf{F}_\theta =\mathbf{F}_\theta(\mathbf{X})$ are given by $\mathbf{Y}^*\equiv 0$ and $\mathbf{U}^*\equiv 0$.
\begin{equation}
    \mathbf{\mathbf{X}^*}=\frac{\Delta t}{\gamma_1}\sigma(\mathbf{F}_\theta(\mathbf{X}^*)), \; \mathbf{\mathbf{H}^*}=\frac{\Delta t}{\gamma_2}\sigma(\mathbf{F}_\theta(\mathbf{H}^*)).
\end{equation}

Thus, this interpretation of \method{} clearly brings out its relationship with equivariant GNNs. Unlike standard multi-layer GNNs of the generic type, which can be thought of as steady states of the underlying ODEs, \method{} dynamically evolves the underlying node features over time. Interpreting the multiple GNN layers as iterations at times $t_n = n\Delta t$, we observe that the node features in \method{} follow the trajectories of the corresponding dynamical system and can explore a richer sampling of the underlying latent feature space, leading to possibly greater expressive power than equivariant GNNs, which might remain in the vicinity of steady states.

{\textbf{Mechanism of Dynamic Evolution in \method{}.} \method{} evolves node features according to a second-order ordinary differential equation, as defined in Eq.~\ref{eq:second_ode}. At each iteration, the feature velocity $U_n$ is first updated in Eq.~\ref{eq:update} based on the previous velocity, the nonlinear transformation of aggregated message passing, and damping terms. The node feature $H_n$ is then updated in Eq.~\ref{ode2} by integrating this velocity. This mechanism allows node features to incorporate both instantaneous message-passing effects and accumulated momentum from previous steps. Consequently, feature evolution in \method{} is determined not only by current neighborhood information, but also by the historical trajectory of feature changes, enabling a smooth and continuous adaptation of node representations throughout the learning process. The coordinates are updated in the same manner, following an analogous second-order dynamic process as described for the node features.}

{\textbf{Intuitive Understanding for Over-smoothing Mitigation of \method{}.} Different first-order GNNs, where repeated aggregation causes node features to become increasingly similar, resulting in over-smoothing, our second-order ODE framework introduces momentum-like effects into the evolution of both features and coordinates. The second-order terms ensure that updates to each node’s feature and coordinate depend not only on their current state and those of their neighbors, but also on the direction and rate of change from previous steps, analogous to acceleration in physical systems~\cite{norcliffe2020second,rusch2022graph}. This added momentum helps preserve the diversity and inertia of node representations and coordinates, preventing rapid or uniform convergence. As a result, the network can be made much deeper without suffering from over-smoothing, as both the features and their geometric arrangement in latent space retain discriminative power across many layers.}

\subsection{{Time Complexity Analysis}}
\label{sec:time}
{
The computational complexity of \method{} is determined by the coupled evolution of both node representations (features) and node positions (coordinates) via second-order ODEs. At each ODE solver step, the framework performs two independent forward passes through the underlying equivariant GNN backbone—one for the feature branch and one for the coordinate branch.}

{
Let $N$ denote the number of nodes, $d$ the average node degree, $d_f$ the hidden feature dimension, and $T$ the number of ODE solver steps. The overall time complexity per iteration depends on the choice of backbone:
\begin{itemize}
    \item \textbf{DuSEGO-EGNN:} For the EGNN~\cite{satorras2021n} backbone, each message passing layer involves lightweight computations such as MLPs and pairwise distance calculations. The complexity of a single forward pass is $\mathcal{O}(N \cdot d \cdot d_f)$. Since both node features and coordinates are updated independently, the total complexity for $T$ solver steps is:
    \begin{equation*}
        \mathcal{O}(2T \cdot N \cdot d \cdot d_f).
    \end{equation*}
    \item \textbf{DuSEGO-SEGNN:} The SEGNN~\cite{brandstetter2021geometric} backbone incorporates O(3)-steerable message passing, which requires higher-order tensor products and equivariant nonlinearities. The computational cost per forward pass is approximately $\mathcal{O}(N \cdot d \cdot d_f^3)$. Thus, the total complexity for both branches is:
    \begin{equation*}
        \mathcal{O}(2T \cdot N \cdot d \cdot d_f^3).
    \end{equation*}
\end{itemize}
}

\section{\method{} Properties Analysis}

To gain some insights into the functioning of \method{}, we start by setting the hyperparameter $\gamma=1$ and assuming that the 1-neighborhood coupling $F_\theta$ is given by either the GAT or GCN type coupling functions. In this case, the underlying ODEs take the following node-wise form, 

\begin{proposition}
    Suppose the backbone Equivariant GNN $F_\theta$ of \method{} is \textit{E}(3) equivariant and translation- invariant, the output $\mathbf{H}$ and $\mathbf{X}$ is \textit{E}(3)-equivariant.
\end{proposition}

\textit{Proof:}
Due to $F_\theta$ is \textit{E}(3) equivariant, then we have:
\begin{equation}
    F_\theta(\mathbf{X}+g, \mathbf{H}+g) = F_\theta(\mathbf{X}, \mathbf{H})+g,\quad F_\theta(Q\mathbf{X}, Q\mathbf{H}) = QF_\theta(\mathbf{X}, \mathbf{H}).
\end{equation}

For the proposed \method{}, we ignore the $\sigma$ and set $\gamma_1=1$, $\gamma_2=1$ for simplification,
\begin{equation}
    F_\theta(\mathbf{X}+g,\mathbf{H}+g)-(\mathbf{X}+g)-\alpha((\mathbf{X}+g)')=F_\theta(\mathbf{X},\mathbf{H})+g-\mathbf{X}-g-\alpha \mathbf{X}'=F_\theta(\mathbf{X},\mathbf{H})-\mathbf{X}-\alpha \mathbf{X}'=\mathbf{X}''\nonumber,
\end{equation}
\begin{equation}
    F_\theta(\mathbf{X}+g,\mathbf{H}+g)-(\mathbf{H}+g)-\alpha((\mathbf{H}+g)')=F_\theta(\mathbf{X},\mathbf{H})+g-\mathbf{H}-g-\alpha \mathbf{H}'=F_\theta(\mathbf{X},\mathbf{H})-\mathbf{H}-\alpha \mathbf{H}'=\mathbf{H}''\nonumber.
\end{equation}

Similarly, we have:
\begin{equation}
    F_\theta(Q\mathbf{X},Q\mathbf{H})-Q\mathbf{X}-\alpha Q\mathbf{X}'=QF_\theta(\mathbf{X},\mathbf{H})-Q\mathbf{X}-\alpha Q\mathbf{X}'=Q\mathbf{X}'',
\end{equation}
\begin{equation}
    F_\theta(Q\mathbf{X},Q\mathbf{H})-Q\mathbf{H}-\alpha Q\mathbf{H}'=QF_\theta(\mathbf{X},\mathbf{H})-Q\mathbf{H}-\alpha Q\mathbf{H}'=Q\mathbf{H}''.
\end{equation}
Therefore, the output $\mathbf{H}$ and $\mathbf{X}$ also follows \textit{E}(3) equivariant.

\begin{proposition}
    Let $\mathbf{X}_n$, $\mathbf{Y}_n$ be the node features, and $\mathbf{H}_n$, $\mathbf{U}_n$ be the node coordinate, generated by \method{}. We assume that $\Delta t \ll 1$ is chosen to be sufficiently small. Then, the gradient of the loss function $\mathcal{L}$ with respect to any learnable weight parameter $\mathbf{w}_k^l$, for some $1\leq k\leq v$ and $1\leq l\leq N$ is bounded as:
    \begin{equation}
\begin{aligned}
    \left|\frac{\partial \mathcal{L}}{\partial \mathbf{w}_k^l} \right|\leq& \frac{\hat{D}\Delta t}{v}\Bigg(\max\limits_{x}|\sigma'(x)|\cdot(1+\Gamma_x N\Delta t)\Big(\max\limits_{1\leq i\leq v}|\mathbf{X}_i^0|+|\mathbf{Y}_i^0|\Big)\\
    &+\max\limits_{h}|\sigma'(h)|\cdot(1+\Gamma_h N\Delta t)\Big(\max\limits_{1\leq i\leq v}|\mathbf{H}_i^0|+|\mathbf{U}_i^0|\Big)\Bigg)\\
    &+\frac{\hat{D}\Delta t}{v}\Bigg(\max\limits_{x}|\sigma'(x)|\cdot(1+\Gamma_x N\Delta t)\Big(\max\limits_{1\leq i\leq v}|\bar{\mathbf{X}}_i|+\max\limits_x|\sigma(x)|\sqrt{N\Delta t}\Big)^2\\
    &+\max\limits_{h}|\sigma'(h)|\cdot(1+\Gamma_h N\Delta t)\Big(\max\limits_{1\leq i\leq v}|\bar{\mathbf{H}}_i|+\max\limits_h|\sigma(h)|\sqrt{N\Delta t}\Big)^2\Bigg).
\end{aligned}
\end{equation}
    where 
    $\hat{D}=\max\limits_{i,j\in\mathcal{V}}\frac{1}{\sqrt{d_id_j}}$, and $\Gamma_x:=6+4\max\limits_x|\sigma^{'}(x)|\cdot\hat{D}\max\limits_{1\leq n\leq T}||\mathbf{w}^n||_1$ and $\Gamma_h:=6+4\max\limits_h|\sigma^{'}(h)|\cdot\hat{D}\max\limits_{1\leq n\leq T}||\mathbf{w}^n||_1$. $d_i$ is the degree of node $i$, $\mathbf{\bar{X}}_i$ is the label of node $i$ and $\mathbf{\bar{H}}_i$ is the coordinate of node $i$. 
\end{proposition}

\textit{Proof:}

From~\cite{rusch2022graph}, we have:
\begin{equation}
    \left\Vert \frac{\partial \mathbf{H^N}}{\partial \mathbf{H}^l} \right\Vert_\infty \leq 1+(N-l)\Gamma_h \Delta t\leq 1+N\Gamma_h \Delta t, \quad \left\Vert \frac{\partial \mathbf{X^N}}{\partial \mathbf{X}^l} \right\Vert_\infty \leq 1+(N-l)\Gamma_x \Delta t\leq 1+N\Gamma_x \Delta t,
    \label{eq13}
\end{equation}
where $\Gamma_h=6+4 \max \limits_h |\sigma'(h)|\hat{D}\max\limits_{1\leq n\leq T} \Vert \mathbf{w}^n \Vert_1$, $\Gamma_x=6+4 \max \limits_x |\sigma'(x)|\hat{D}\max\limits_{1\leq n\leq T} \Vert \mathbf{w}^n \Vert_1$, and $\hat{D}=\max\limits_{i,j\in\mathcal{V}}\frac{1}{\sqrt{d_id_j}}$. By the chain rule, the gradient of $\mathbf{w}_k^l$ is:
\begin{equation}
    \frac{\partial \mathcal{L}}{\partial \mathbf{w}_k^l}=\frac{\partial \mathcal{L}}{\partial \mathbf{H}^N} \frac{\partial \mathbf{H}^N}{\partial \mathbf{H}^l} \frac{\partial \mathbf{H}^l}{\partial \mathbf{w}_k^l}+\frac{\partial \mathcal{L}}{\partial \mathbf{X}^N} \frac{\partial \mathbf{X}^N}{\partial \mathbf{X}^l} \frac{\partial \mathbf{X}^l}{\partial \mathbf{w}_k^l},
\end{equation}
where $\frac{\partial \mathbf{H}^N}{\partial \mathbf{H}^l}=\Pi^N_{n=l+1} \frac{\partial \mathbf{H}^n}{\partial \mathbf{H}^{n-1}}$, $\frac{\partial \mathbf{X}^N}{\partial \mathbf{X}^l}=\Pi^N_{n=l+1} \frac{\partial \mathbf{X}^n}{\partial \mathbf{X}^{n-1}}$. With Eq. 51 in~\cite{rusch2022graph}, we directly have:
\begin{equation}
    \left\Vert \frac{\partial \mathcal{L}}{\partial \mathbf{X}^N} \right\Vert_\infty \leq \frac{1}{v}\left(\max\limits_{1\leq i\leq v}(|\mathbf{X}_i^0|+|\mathbf{Y}_i^0|)+\max\limits_{1\leq i\leq v}|\bar{\mathbf{X}}_i|+\max\limits_x|\sigma(x)|\sqrt{N\Delta t}\right),
    \label{eq15}
\end{equation}
\begin{equation}
    \left\Vert \frac{\partial \mathcal{L}}{\partial \mathbf{H}^N} \right\Vert_\infty \leq \frac{1}{v}\left(\max\limits_{1\leq i\leq v}(|\mathbf{H}_i^0|+|\mathbf{U}_i^0|)+\max\limits_{1\leq i\leq v}|\bar{\mathbf{H}}_i|+\max\limits_h|\sigma(h)|\sqrt{N\Delta t}\right),
    \label{eq16}
\end{equation}
where $\bar{\mathbf{X}}$ and $\bar{\mathbf{H}}$ are the ground truth vector.
Similarly, we can readily calculate that:
\begin{equation}
    \left\Vert \frac{\partial \mathbf{X}^l}{\partial \mathbf{w}_k^l}\right\Vert_\infty \leq \Delta t \max\limits_{x}|\sigma'(x)|\hat{D}\left(\max\limits_{1\leq i\leq v}(|\mathbf{X}_i^0|+|\mathbf{Y}_i^0|)+\max\limits_{1\leq i\leq v}|\bar{\mathbf{X}}_i|+\max\limits_x|\sigma(x)|\sqrt{N\Delta t}\right),
    \label{eq17}
\end{equation}
\begin{equation}
    \left\Vert \frac{\partial \mathbf{H}^l}{\partial \mathbf{w}_k^l}\right\Vert_\infty \leq \Delta t \max\limits_{h}|\sigma'(h)|\hat{D}\left(\max\limits_{1\leq i\leq v}(|\mathbf{H}_i^0|+|\mathbf{U}_i^0|)+\max\limits_{1\leq i\leq v}|\bar{\mathbf{H}}_i|+\max\limits_h|\sigma(h)|\sqrt{N\Delta t}\right).
    \label{eq18}
\end{equation}

Therefore, by multiplying Eq.~\ref{eq13}, ~\ref{eq15}-\ref{eq18}, we have:
\begin{equation}
\begin{aligned}
    \left|\frac{\partial \mathcal{L}}{\partial \mathbf{w}_k^l} \right|\leq& \frac{\hat{D}\Delta t}{v}\Bigg(\max\limits_{x}|\sigma'(x)|\cdot(1+\Gamma_x N\Delta t)\Big(\max\limits_{1\leq i\leq v}|\mathbf{X}_i^0|+|\mathbf{Y}_i^0|\Big)\\
    &+\max\limits_{h}|\sigma'(h)|\cdot(1+\Gamma_h N\Delta t)\Big(\max\limits_{1\leq i\leq v}|\mathbf{H}_i^0|+|\mathbf{U}_i^0|\Big)\Bigg)\\
    &+\frac{\hat{D}\Delta t}{v}\Bigg(\max\limits_{x}|\sigma'(x)|\cdot(1+\Gamma_x N\Delta t)\Big(\max\limits_{1\leq i\leq v}|\bar{\mathbf{X}}_i|+\max\limits_x|\sigma(x)|\sqrt{N\Delta t}\Big)^2\\
    &+\max\limits_{h}|\sigma'(h)|\cdot(1+\Gamma_h N\Delta t)\Big(\max\limits_{1\leq i\leq v}|\bar{\mathbf{H}}_i|+\max\limits_h|\sigma(h)|\sqrt{N\Delta t}\Big)^2\Bigg).
\end{aligned}
\end{equation}

\begin{table}[t]
\centering
\tabcolsep=10pt
\setlength{\abovecaptionskip}{0.2cm}
\caption{Mean squared error $\left(\times 10^{-2}\right)$ for the future coordinate estimation under different timesteps in the N-body system experiment, and forward time in seconds for a batch size of 100 samples running on NVIDIA A100 GPUs. Results averaged across five runs. We report both the mean and the standard deviation of MSE in the table. The best performance is bold.}
\centering
\label{table_1}

\resizebox{0.8\textwidth}{!}{
\begin{tabular}{lcccc}%
\toprule
\multirow{2}{*}{Methods}  & \multicolumn{3}{c}{MSE}& \multirowcell{2}{Forward\\Time/ms}  \\
\cmidrule(lr){2-4}
&1000 ts & 1500 ts & 2000 ts   \\
\midrule
Linear & 6.831 $\pm$ 0.016 & 20.012 $\pm$ 0.029 & 39.513 $\pm$ 0.061 & 0.1 \\ 
SE(3) Trans & 2.483 $\pm$ 0.099 & 18.891 $\pm$ 0.287 & 36.730 $\pm$ 0.381 & 114.6 \\
TFN & 1.544 $\pm$ 0.231 & 11.116 $\pm$ 2.825 & 23.823 $\pm$ 3.048 & 27.2 \\
GNN & 1.077 $\pm$ 0.004 & 5.059 $\pm$ 0.250 & 10.591 $\pm$ 0.352 & 2.1 \\
Radial Field & 1.060 $\pm$ 0.007 & 12.514 $\pm$ 0.089 & 26.388 $\pm$ 0.331 & 3.0 \\
EGNN & \space 0.716 $\pm$ 0.029  & 2.201 $\pm$ 0.081 & 4.049 $\pm$ 0.103  & 5.5 \\
SEGNN & \space 0.481 $\pm$ 0.016  & 1.552 $\pm$ 0.061 & 3.294 $\pm$ 0.095 & 24.5 \\ 
\midrule
GraphCon & 1.015 $\pm$ 0.011 & 3.148 $\pm$ 0.091 & 6.133 $\pm$ 0.143 & 10.8 \\ 
SEGNO & \space 0.976 $\pm$ 0.012  & 2.994 $\pm$ 0.059 & 7.453 $\pm$ 0.181 & 21.2 \\
GMN & \space 0.701 $\pm$ 0.018 & 1.956 $\pm$ 0.096 & 3.939 $\pm$ 0.151 & 6.2 \\
\method{}-EGNN & \space 0.674 $\pm$ 0.014 & 1.912 $\pm$ 0.055 & 3.876 $\pm$ 0.079 & 13.9 \\
\method{}-SEGNN &  \space \textbf{0.449 $\pm$ 0.012} & \textbf{1.427 $\pm$ 0.047} & \textbf{3.122 $\pm$ 0.081} & 45.7 \\
\midrule
\end{tabular}
}
\end{table}

\section{Experiments}

To verify the effectiveness of \method{} in enhancing the performance of Equivariant GNN, we follow the experimental settings in \cite{satorras2021n} and evaluate it on three different tasks, i.e., modeling a dynamical system, learning unsupervised representations of graphs in a continuous latent space, and predicting molecular chemical properties. 

\subsection{Modeling a Dynamic System}
In this section, we evaluate \method{} on one simulated N-body system named charged particles, where we aim to forecast the coordinate of particles across different time steps \cite{satorras2021n}.

\subsubsection{Dataset} 

Building on the work of \cite{fuchs2020se}, \cite{satorras2021n} extended the Charged Particles N-body experiment introduced in \cite{kipf2018neural}, where the system consists of five particles, each defined by charge, coordinate, and velocity in three-dimensional space. In this setup, particles in each graph are fully connected, with the initial velocity norm included as an additional node feature. Using the N-body system code from \cite{satorras2021n}, we generated a new dataset for the charged N-body system across 1,000, 1,500, and 2,000 timesteps. The initial coordinates of particles were sampled from a Gaussian distribution with $\mu=0$ and $\sigma=0.5$, and the initial velocity was represented as a random vector with a norm of 0.5. The dataset is split into training, validation, and testing sets, containing 10,000, 2,000, and 2,000 samples, respectively.

\begin{figure}[t]
\centering
\begin{subfigure}[b]{0.3\textwidth}
    \includegraphics[width=\textwidth]{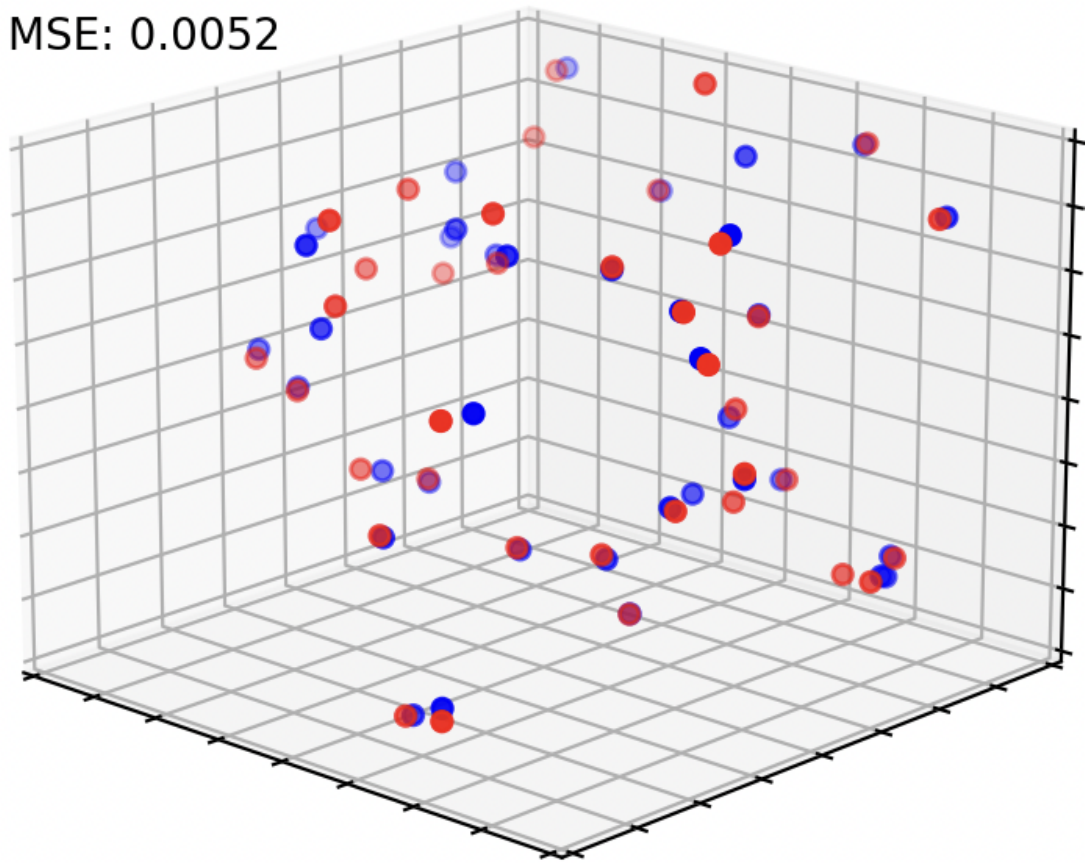}
    \caption{GMN}
    \label{fig:image1}
\end{subfigure}
\hfill
\begin{subfigure}[b]{0.3\textwidth}
    \includegraphics[width=\textwidth]{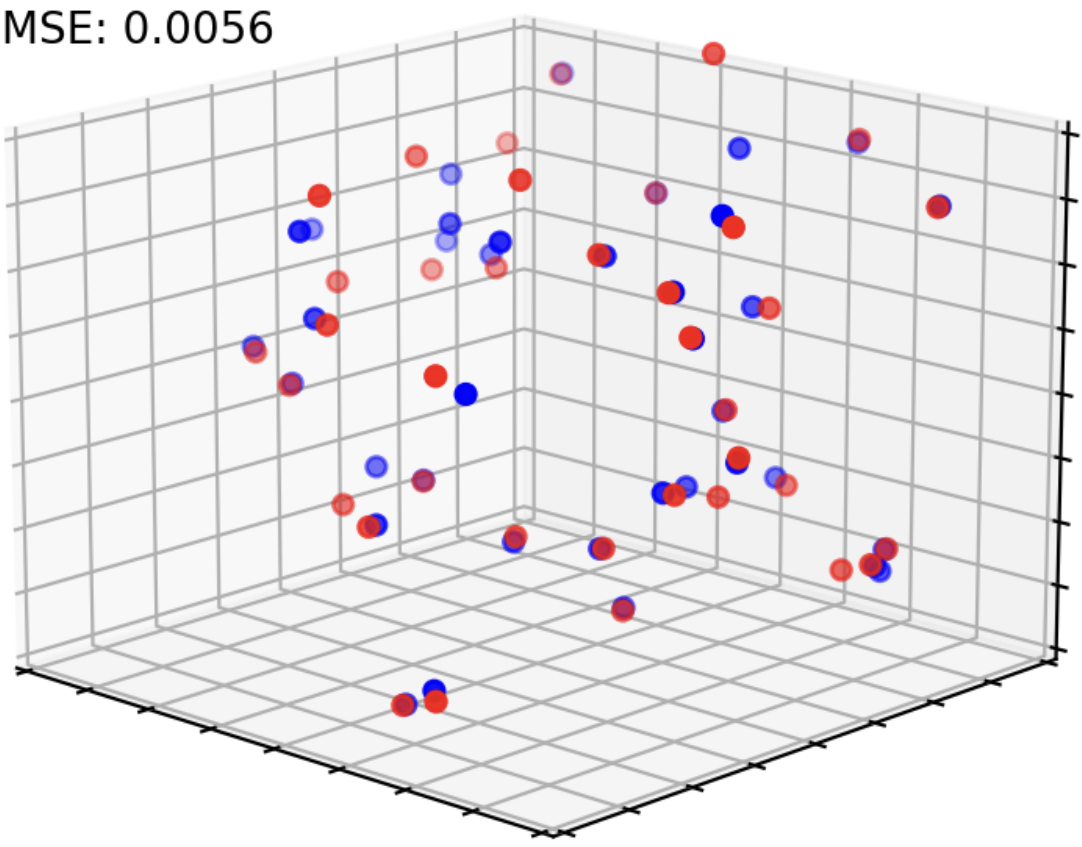}
    \caption{EGNN}
    \label{fig:image2}
\end{subfigure}
\hfill
\begin{subfigure}[b]{0.3\textwidth}
    \includegraphics[width=\textwidth]{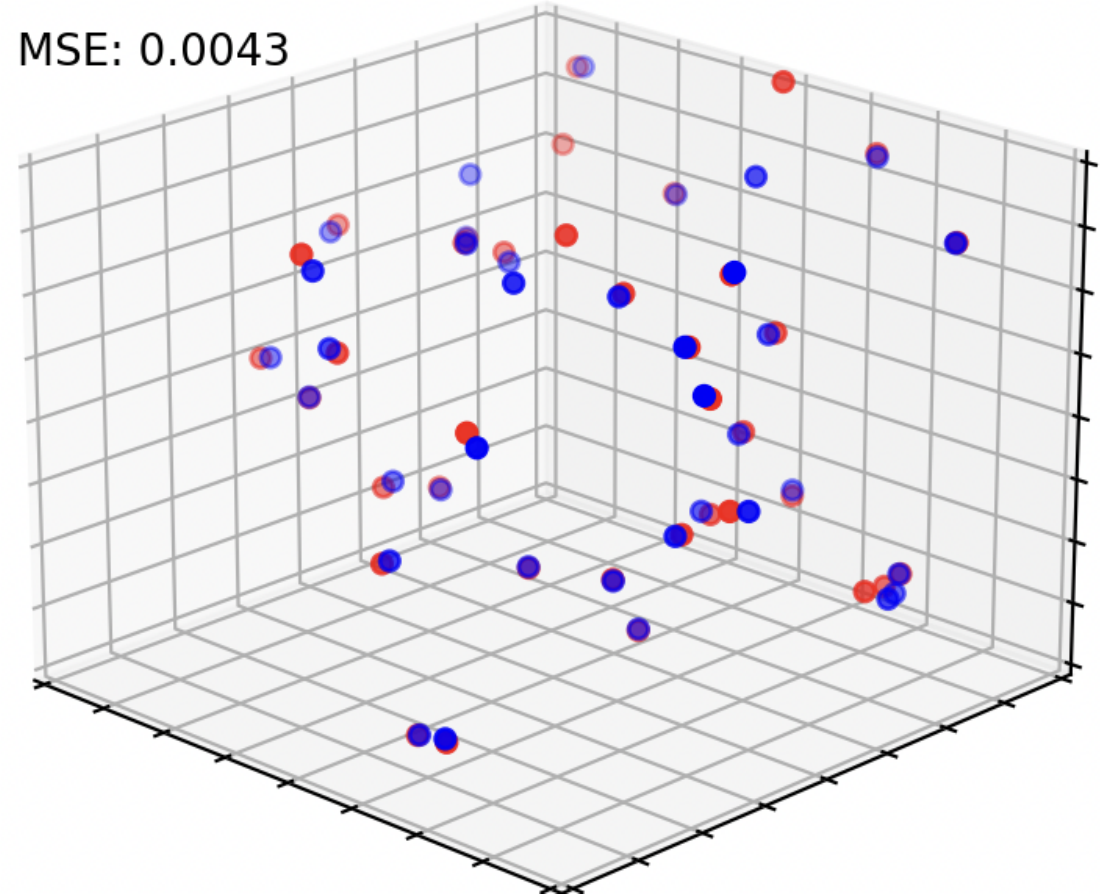}
    \caption{SEGNN}
    \label{fig:image3}
\end{subfigure}

\begin{subfigure}[b]{0.3\textwidth}
    \includegraphics[width=\textwidth]{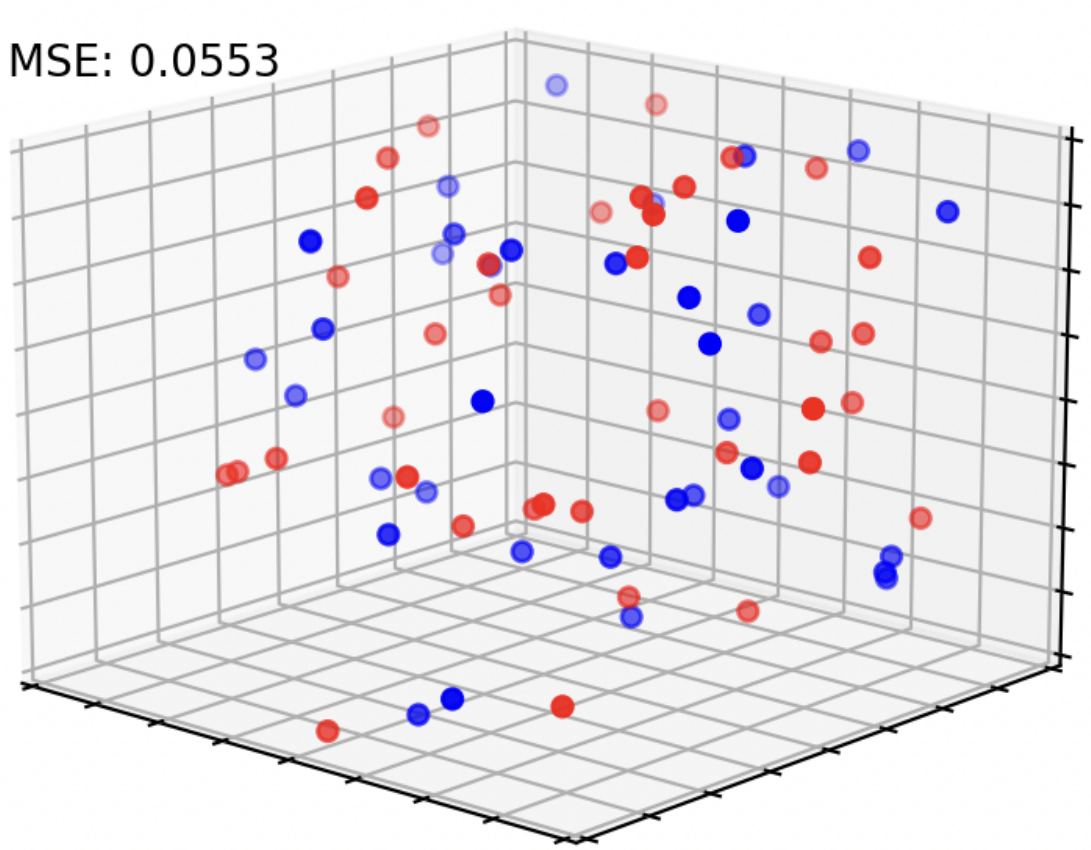}
    \caption{SEGNO}
    \label{fig:image4}
\end{subfigure}
\hfill
\begin{subfigure}[b]{0.3\textwidth}
    \includegraphics[width=\textwidth]{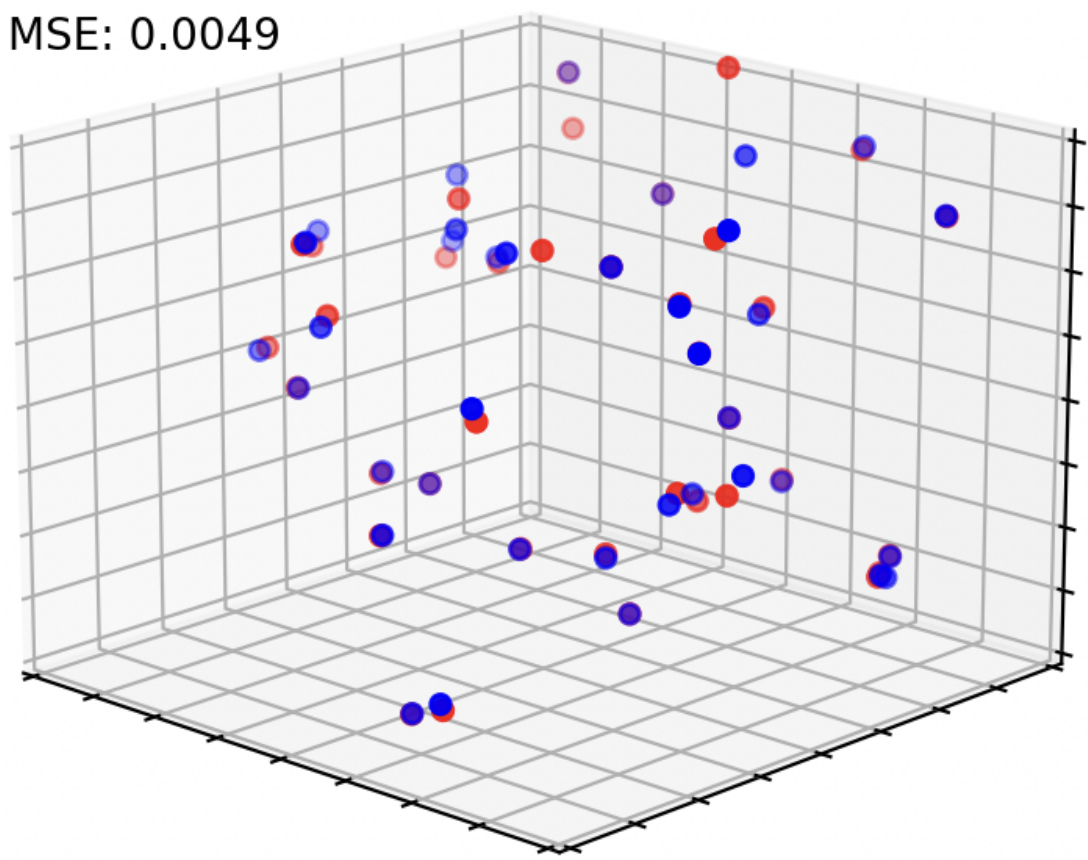}
    \caption{\method{}-EGNN }
    \label{fig:image5}
\end{subfigure}
\hfill
\begin{subfigure}[b]{0.3\textwidth}
    \includegraphics[width=\textwidth]{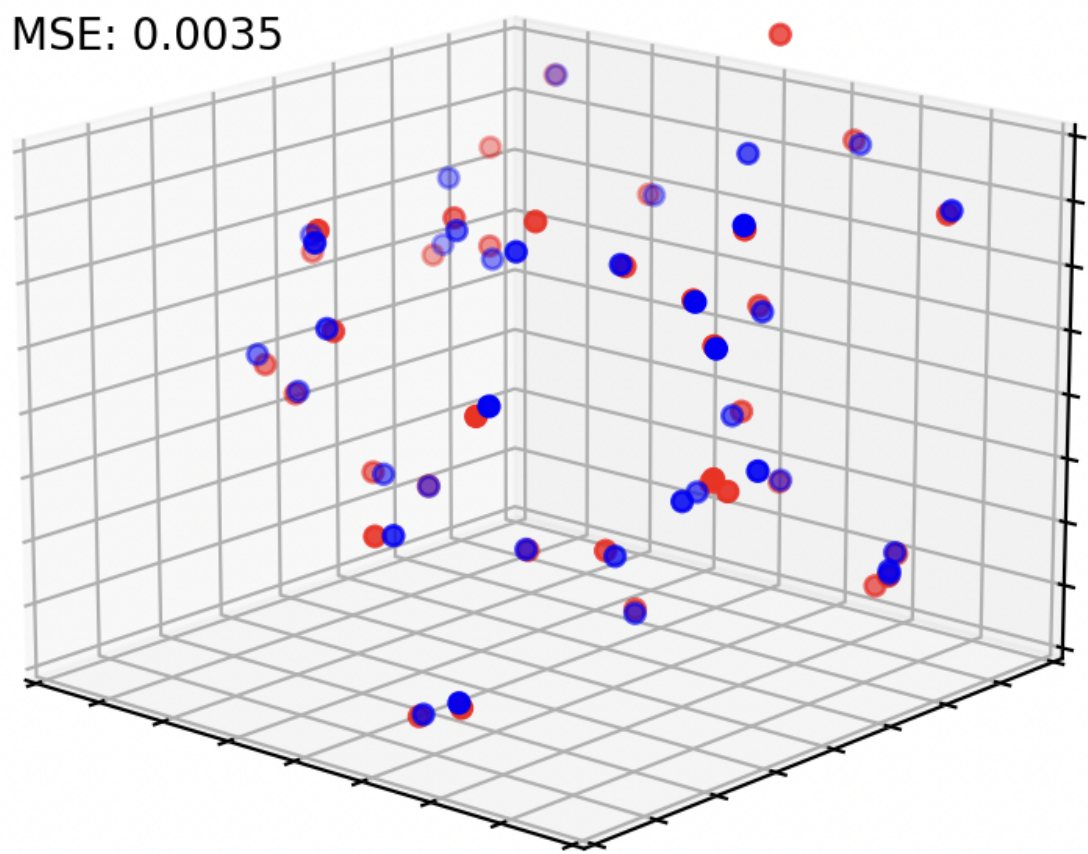}
    \caption{\method{}-SEGNN }
    \label{fig:image6}
\end{subfigure}
\caption{Visualization of ground truths (\textcolor{blue}{Blue}) and prediction results (\textcolor{red}{Red}) for EGNN, SEGNN, GMN, SEGNO, \method{}-EGNN, and \method{}-SEGNN under 1000 timesteps on the N-body system dataset.}
\label{fig:figure_vis}
\vspace{-0.5cm}
\end{figure}

\subsubsection{Baselines.} Two types of baseline methods are deployed for fair comparisons: (1) Methods that incorporate first-order information, including the Linear model, Graph Neural Networks (GNN), SE(3) Transformer \cite{fuchs2020se}, Tensor Field Networks (TFN) \cite{thomas2018tensor}, Radial Field (RF) \cite{kohler2019equivariant}, Equivariant Graph Neural Networks (EGNN) \cite{satorras2021n}, and Steerable E(3) Equivariant Graph Neural Networks (SEGNN) \cite{brandstetter2021geometric}; (2) Models that incorporate second-order information, including GraphCon \cite{rusch2022graph}, Graph Mechanics Network (GMN) \cite{huang2021equivariant}, and Second-order Equivariant Graph Neural Ordinary Differential Equation (SEGNO) \cite{liu2024segno}. 

\subsubsection{Experimental Setting.} We build on the experimental framework introduced in \cite{liu2024segno}, aiming to estimate the coordinates of all particles after 1,000, 1,500, and 2,000 timesteps. We re-evaluate all baseline methods under identical experimental settings on NVIDIA A100 GPUs, gathering results across five runs to ensure fair comparisons. The experimental setup is as follows: Adam optimizer with a learning rate of $5 \times 10^{-4}$, batch size of 100, 10,000 epochs, hidden embedding dimension of 64, weight decay of $1 \times 10^{-12}$, and four GNN layers. Specifically, for the GMN method, we set the numbers of isolated particles, sticks, and hinges to $(5, 0, 0)$. For the SE(3)-transformer and TFN, we set the representation degrees to 3 and 2, respectively. An early stopping criterion is applied, halting training if no improvement in validation loss is observed within 50 epochs.

\subsubsection{Implementation details.} We employ EGNN with the extension that includes velocity and SEGNN as the backbone of \method{}, named \method{}-EGNN and \method{}-SEGNN. Especially, EGNN and SEGNN are implemented based on the implementation details provided in \cite{satorras2021n, brandstetter2021geometric}, which are used as the coupling function for message passing \cite{rusch2022graph}. Then, we use the ODE solvers provided in torchdiffeq \footnote{https://github.com/rtqichen/torchdiffeq} library for solving Eq \ref{ode1} and Eq. \ref{ode2}. Additionally, we provide the forward time in seconds for each method for a batch of 100 samples.

\begin{figure}[t]
\centerline{\includegraphics[width=0.7\textwidth]{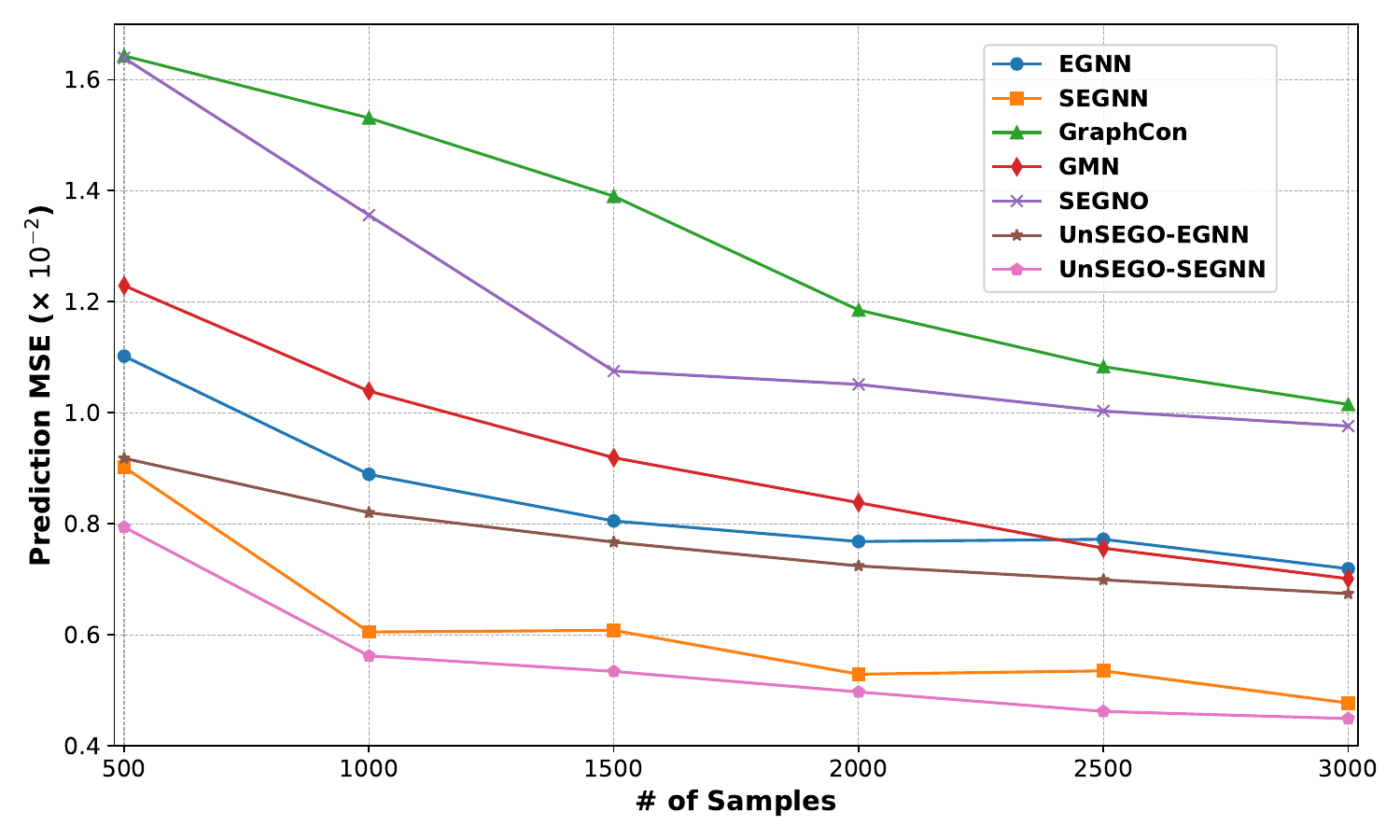}}
    \caption{Data efﬁciency comparisons among EGNN, SEGNN, GMN, GraphCon, SEGNO, \method{}-EGNN , and \method{}-SEGNN  on the N-body dataset.}
    \label{fig:data efficientcy}
    \vspace{-0.5cm}
\end{figure}

\begin{figure}[t]
    \centering
    \begin{subfigure}[b]{0.48\textwidth} 
        \centering
        \includegraphics[width=\textwidth]{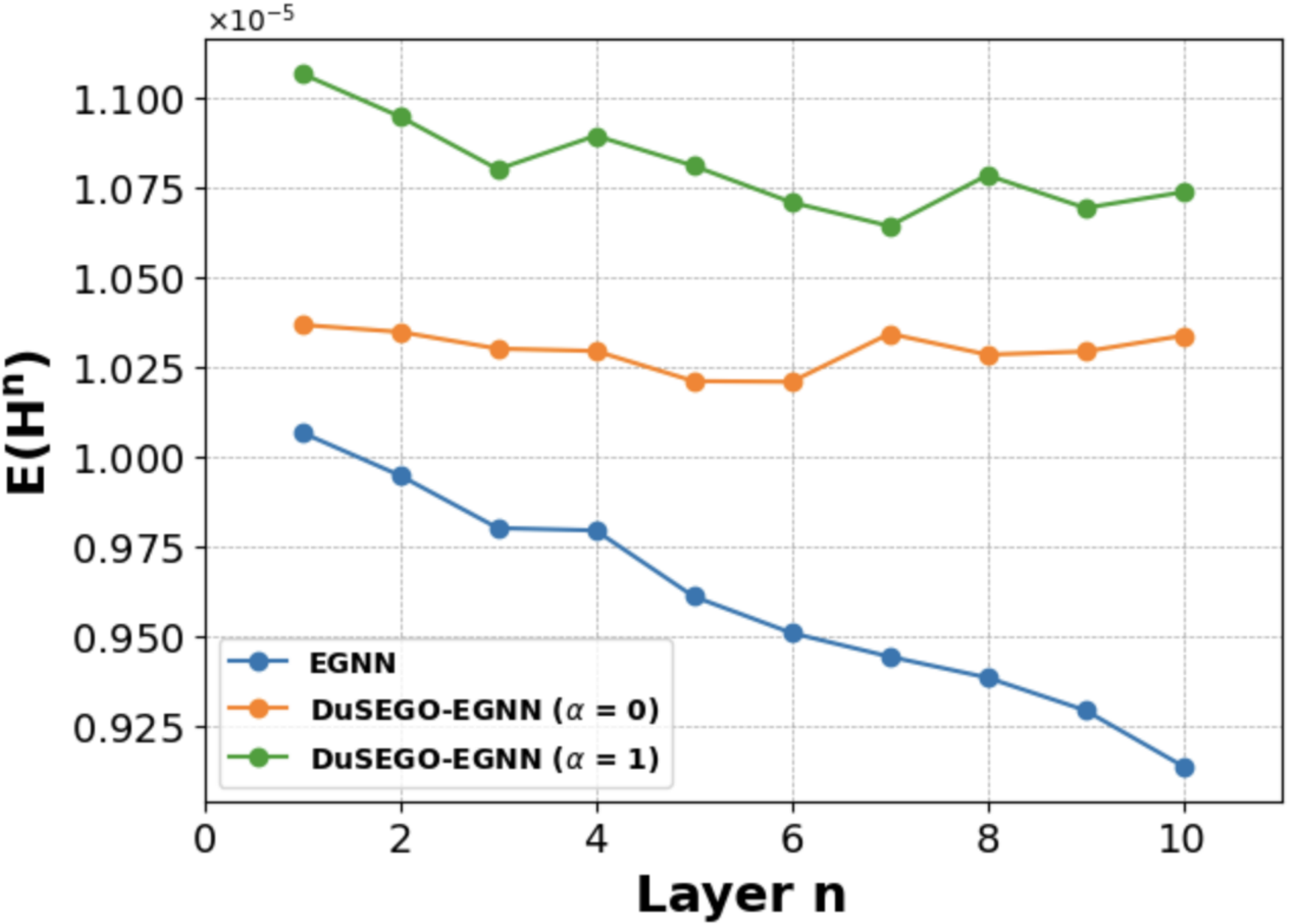}
        \caption{Dirichlet energy $\mathbf{E}\left(\mathbf{H}^n\right)$ of node features.}
        \label{fig:sub3}
    \end{subfigure}
    \hfill 
    \begin{subfigure}[b]{0.48\textwidth}
        \centering
        \includegraphics[width=0.98\textwidth,height=0.714\textwidth]{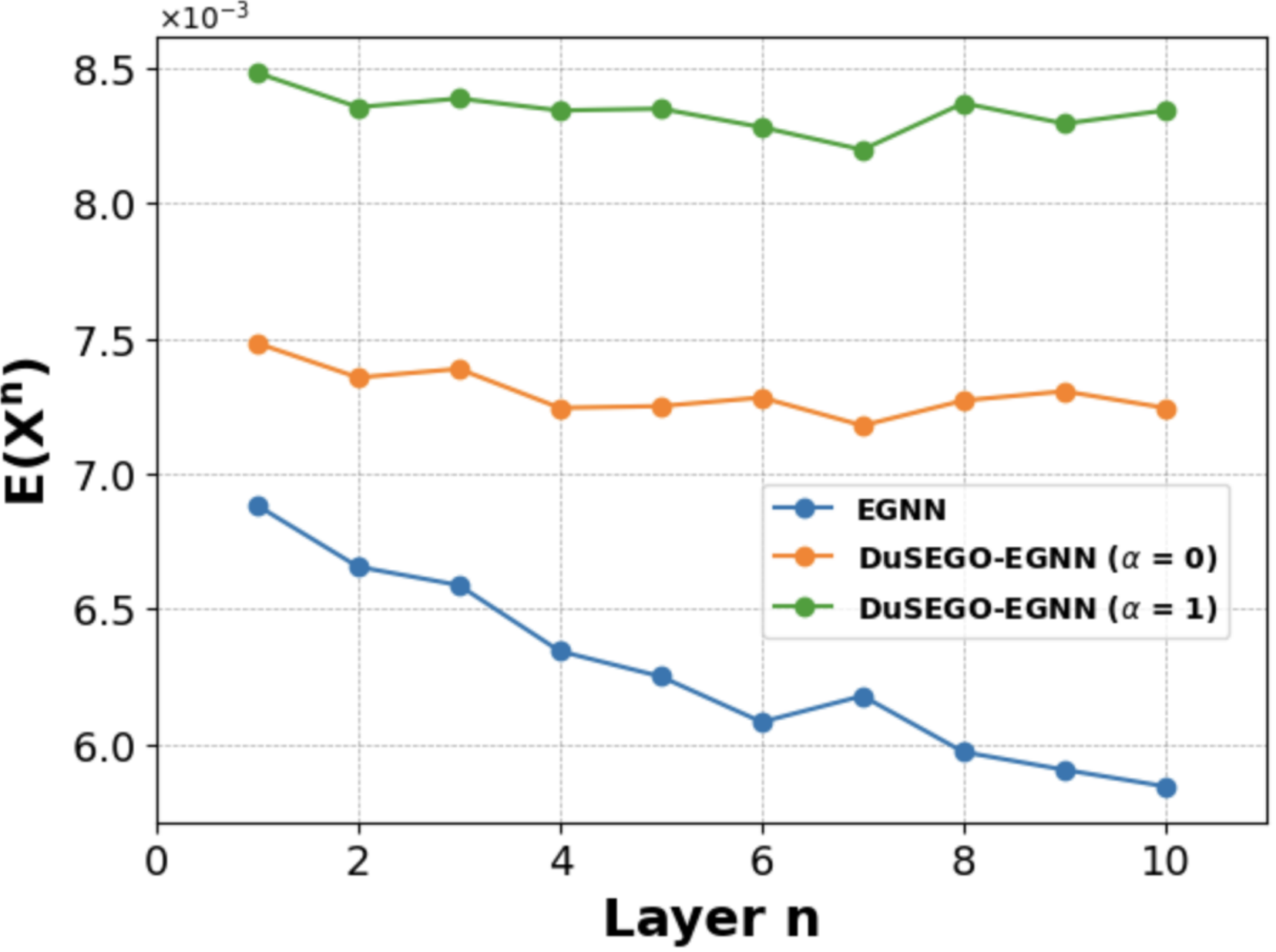}
        \caption{Dirichlet energy $\mathbf{E}\left(\mathbf{X}^n\right)$ of node coordinates.}
        \label{fig:sub4}
    \end{subfigure}
    \caption{Dirichlet energy $\mathbf{E}\left(\mathbf{H}^n\right)$ of node features $\mathbf{H}^n$ and $\mathbf{E}\left(\mathbf{X}^n\right)$ of node coordinates $\mathbf{X}^n$ propagated through a EGNN and \method{}-EGNN for two different values of $\alpha=0,1$ and $\gamma_1 = \gamma_2=1$ in Eq. \ref{ode1} and Eq. \ref{ode2}.}
    \vspace{-0.5cm}
    \label{fig:oversmoothing_egnn}
\end{figure}

\subsubsection{Results.} 
The results for all baseline methods, \method{}-EGNN, and \method{}-SEGNN on the N-body system dataset are shown in Table \ref{table_1}. We observe that \method{} significantly enhances the performance of EGNN and SEGNN, with \method{}-SEGNN outperforming all other baseline methods and \method{}-EGNN under the same experimental conditions. Specifically, \method{}-EGNN and \method{}-SEGNN reduce the mean squared error (MSE) of EGNN and SEGNN by an average of 7.73\% and 6.65\%, respectively, across the three timesteps, indicating substantial improvement. Notably, the relative improvement for \method{}-EGNN increases from 5.86\% at 1,000 timesteps to 13.13\% at 1,500 timesteps, highlighting the strong generalization ability of \method{}. Additionally, Fig. \ref{fig:figure_vis} presents a visualization of GMN, EGNN, SEGNN, GMN, SEGNO, \method{}-EGNN, and \method{}-SEGNN in estimating particle coordinates for the N-body system experiment at 1,000 timesteps, with MSE values between predictions and ground truths reported. This visualization further shows that \method{}-SEGNN and \method{}-EGNN outperform EGNN and SEGNN, respectively, aligning with the performance results in Table \ref{table_1}.

It is crucial to verify that the physical principles identified from a limited number of experiments are sufficiently general to explain phenomena on a broader scale. To this end, we assess the performance of EGNN, SEGNN, GMN, GraphCon, SEGNO, \method{}-EGNN, and \method{}-SEGNN across various training set sizes, ranging from 500 to 3,000 samples in the N-body system dataset. In this evaluation, the timesteps of particle coordinates are set to 1,000. As shown in Fig. \ref{fig:data efficientcy}, \method{}-EGNN and \method{}-SEGNN, similar to the GMN method, demonstrate more stable performance than other baseline methods, emphasizing their data efficiency. This result suggests that data efficiency, a notable strength of \method{}, is essential for effectively learning to model physical systems.

\begin{wraptable}{r}{0.4\linewidth}
  \centering
  \vspace{-1.4em}
  \caption{Training time (in seconds) per epoch for \method{}-EGNN and \method{}-SEGNN on the N-body system with 1000 timesteps under different layers.}
  \label{tab:runtime}
  \resizebox{\linewidth}{!}{
    \begin{tabular}{c|c|c}
      \toprule
      \# Layers & \method{}-EGNN & \method{}-SEGNN \\
      \midrule
      4  &  1.0433  &  4.0484  \\
      8  &  1.9295  &  7.7215  \\
      12 &  2.7909  &  11.7355  \\ 
      16 &  3.6658  &  15.5262  \\
      \bottomrule
    \end{tabular}
  }
\end{wraptable}

To further investigate the over-smoothing issue in EGNN, we implement EGNN and \method{}-EGNN with two different values of $\alpha = 0, 1$ and set $\gamma_1 = \gamma_2 = 1$ in Eq. \ref{ode1} and Eq. \ref{ode2} on the N-body system dataset at 1,000 timesteps. As shown in Fig. \ref{fig:oversmoothing_egnn}, the over-smoothing problem in EGNN becomes apparent as the number of EGNN layers increases. The differences between node features, such as $\mathbf{h}_i^n$ and $\mathbf{h}_j^n$, as well as node coordinates, like $\mathbf{x}_i^n$ and $\mathbf{x}_j^n$, gradually decrease, indicating that node features and coordinates are converging toward a constant value. However, with the integration of \method{} into EGNN, the over-smoothing issue is effectively mitigated, as $\mathbf{E}\left(\mathbf{H}^n\right)$ and $\mathbf{E}\left(\mathbf{X}^n\right)$ remain stable within a reasonable range, even as the number of EGNN layers increases.

Additionally, we compare the performance of EGNN and \method{}-EGNN as the number of EGNN layers increases (up to 16). As shown in Table \ref{tab:table_6}, the MSE for EGNN decreases from layer 4 to layer 8. However, at 12 layers, the gradient vanishing issue emerges, preventing EGNN from producing accurate results. This confirms that gradient vanishing becomes a limitation as the depth of EGNN increases. In contrast, the MSE for \method{}-EGNN continues to decrease with additional layers across different timesteps, and \method{}-EGNN maintains stable performance, supporting our hypothesis that integrating EGNN within \method{} helps mitigate the gradient vanishing problem. We also measured the training time per epoch for \method{}-EGNN and \method{}-SEGNN as the number of GNN layers increased, with timesteps fixed at 1000. As shown in Table~\ref{tab:runtime}, training time grows nearly linearly with network layers, consistent with our theoretical analysis in Section~\ref{sec:time}. Notably, \method{}-SEGNN shows much higher computational costs per layer than \method{}-EGNN, due to the intensive tensor operations in the SEGNN architecture. These results confirm both the scalability and the strong dependence of runtime on the chosen backbone.

\begin{table}[t]
\centering
\tabcolsep=12.5pt
\setlength{\abovecaptionskip}{0.2cm}
\caption{Mean squared error $\left(\times 10^{-3}\right)$ for the future coordinate estimation under different timesteps in the N-body system experiment between EGNN and \method{}-EGNN under different layers. Results averaged across 5 runs.}

\centering
\label{tab:table_6}
\resizebox{0.8\textwidth}{!}{
\begin{tabular}{lccc}
\toprule
\multirow{2}{*}{Methods}  & \multicolumn{3}{c}{MSE}  \\
\cmidrule(lr){2-4}
&1000 ts & 1500 ts & 2000 ts   \\
\midrule
EGNN-layers=4 & \space 7.16 $\pm$ 0.29  & 22.01 $\pm$ 0.81 & 40.49 $\pm$ 1.03  \\
EGNN-layers=8 & \space 6.82 $\pm$ 0.23  & 21.39 $\pm$ 1.46 & 38.97 $\pm$ 2.77  \\
EGNN-layers=12 & \space NAN  & NAN & NAN  \\
EGNN-layers=16 & \space NAN  &  NAN &  NAN  \\
\midrule
\method{}-EGNN-layers=4  & \space 6.74 $\pm$ 0.14 & 19.12 $\pm$ 0.55 & 38.76 $\pm$ 0.79 \\
\method{}-EGNN-layers=8  & \space 6.68 $\pm$ 0.12 & 18.42 $\pm$ 0.78 & 37.46 $\pm$ 1.13 \\
\method{}-EGNN-layers=12  & \space 6.62 $\pm$ 0.11 & 17.62 $\pm$ 0.68 & 36.57 $\pm$ 1.62 \\
\method{}-EGNN-layers=16  & \space 6.49 $\pm$ 0.18  & 17.03 $\pm$ 0.99 & 35.91 $\pm$ 2.01 \\
\midrule
\end{tabular}
}
\vspace{-0.4cm}
\end{table}

\subsection{Graph Autoencoder}

Subsequently, we compare \method{} with the state-of-art baselines for the Graph Autoencoder task, where we aim to estimate the adjacency matrix ${A}_{ij}$ of a graph $\mathcal{G}$ generated from the graph autoencoder and minimize the binary cross entropy between the estimated adjacency matrix $\hat{A}_{ij}$ and the ground truth $A_{ij}$.

\subsubsection{Dataset.} Two widely used datasets are deployed in this section: (1) \textbf{Community} ~\cite{kipf2016variational} dataset, generated using the Erdos-Renyi model \cite{erdds1959random}, consists of random graphs where nodes are connected with a predefined probability, capturing the diversity in the density and structure of various communities. This model, denoted as $G(n, p)$, generates networks by independently linking pairs of nodes with probability $p$; (2) \textbf{Erdos \& Renyi} \cite{janson2011random} dataset, based on the foundational random graph generation models \cite{janson2011random, you2018graphrnn} and also represented as $G(n, p)$, includes a collection of graphs that exhibit the randomness and variability of network structures.

Following \cite{satorras2021n}, we generate the Community and Erdos \& Renyi datasets \cite{kohler2019equivariant, kipf2018neural} using the code provided in \cite{kipf2018neural}. Specifically, we sample 50,000 graphs for training, 5,000 for validation, and 5,000 for testing, where we set $p=0.1$ and $6 \leq n \leq 10$ for Erdos\&Renyi, and $p=0.15$ and $6 \leq n \leq 11$ for Community, respectively. Additionally, 20 communities are divided in the Community dataset, and each graph is deﬁned as an adjacency matrix $A \in \{0,1\}^{M \times M}$ in these two datasets.

\begin{table}
\tabcolsep=7pt
\caption{\% Error and F1 scores for the test partition on the Graph Autoencoding experiment in the Community and Erdos \& Renyi datasets. Results averaged across 5 runs. The best performance is bold.}
\centering
\label{tab:autoencoder1}
\resizebox{0.55\textwidth}{!}{
\begin{tabular}{lcccc}
\toprule
\multirow{2}{*}{Methods}  & \multicolumn{2}{c}{Community}& \multicolumn{2}{c}{Erdos \& Renyi}  \\
\cmidrule(lr){2-3} \cmidrule(lr){4-5}
& F1 & \% Error & F1 & \% Error   \\
\midrule
GNN & 0.957 & 3.86  &0.776 & 6.44 \\ 
GNN-Noise & 0.971 & 2.54 & 0.919 & 1.79\\ 
RFN & 0.973 & 1.81  &0.921 & 1.45\\
EGNN & 0.981 & 1.58  &0.933 & 1.26 \\
\midrule
GraphCon & 0.960 & 3.79  & 0.873 & 3.77 \\
SEGNO & 0.980  & 1.66  & 0.920  & 1.43 \\ 
GMN & 0.979 & 1.74 &0.918 & 1.46\\ 
\method{}-EGNN & \textbf{0.984} & \textbf{1.48}  &\textbf{0.936} & \textbf{1.18} \\
\bottomrule
\end{tabular}
}
\vspace{-0.1cm}

\end{table}

\begin{table}
\centering
\tabcolsep=12.5pt
\setlength{\abovecaptionskip}{0.2cm}
\caption{\% Error and F1 scores for the test partition on the Graph Autoencoder experiment in the Erdos\&Renyi and Community datasets between EGNN and \method{}-EGNN under different layers. Results averaged across 5 runs. }
\centering
\label{tab:autoencoder2}
\resizebox{0.8\textwidth}{!}{
\begin{tabular}{lcccc}
\toprule
\multirow{2}{*}{Methods}  & \multicolumn{2}{c}{Community}& \multicolumn{2}{c}{Erdos \& Renyi}  \\
\cmidrule(lr){2-3} \cmidrule(lr){4-5}
& F1 & \% Error & F1 & \% Error   \\
\midrule
EGNN-layers=4 & 0.981 & 1.58  & 0.933 & 1.26 \\ 
EGNN-layers=6 & 0.977 & 2.04 & 0.929 & 1.39\\ 
EGNN-layers=8 & Inf & Inf  &0.925 & 1.46 \\
EGNN-layers=12 & Inf & Inf  & Inf & Inf \\
EGNN-layers=16 & Inf & Inf  & Inf & Inf \\
\midrule
\method{}-EGNN-layers=4 & 0.984 & 1.48  &0.936 & 1.18 \\ 
\method{}-EGNN-layers=6 & 0.980 & 1.76 & 0.957 & 0.77\\ 
\method{}-EGNN-layers=8 & 0.981 & 1.62  &0.963 & 0.63\\
\method{}-EGNN-layers=12 & 0.982 & 1.57  &0.959 & 0.79 \\
\method{}-EGNN-layers=16 & 0.982 & 1.58  &0.966 & 0.59 \\
\bottomrule
\end{tabular}
}
\end{table}

\subsubsection{Baselines} Two types of baseline methods are deployed for comparison: (1) Methods that contains first-order information including GNN, GNN-Noise, RF \cite{kohler2019equivariant}, EGNN \cite{satorras2021n}; (2) Methods that contains second-order information including GraphCon \cite{rusch2022graph}, GMN \cite{huang2021equivariant} and SEGNO \cite{liu2024segno}.

\subsubsection{Experimental Setting.} The noise generated from $\mathcal{N}(\mathbf{0}, \sigma \mathbf{I}) \in \mathbb{R}^{M \times n}$ will be used as input for all baseline methods. The output will then be employed to embed the graph and fed into the graph decoder to generate an estimated adjacency matrix, $\hat{A}_{ij}$. We rerun all baseline methods under the same experimental settings on NVIDIA A100 GPUs and collect the results across five runs for a fair comparison. The experimental settings are as follows: Adam optimizer with learning rate ${1 \times 10^{-3}}$, the number of epochs 1,000, batch size 256, hidden embedding dim 64, weight decay ${1 \times 10^{-12}}$, layer number 4, and dimensions for the embedding space 8. Velocity is not considered in the Graph Autoencoder task,  which is set to None in the EGNN, GMN, SEGNO, and \method{}-EGNN. 

\subsubsection{Implementation details.} We build upon the experimental setting introduced in \cite{satorras2021n} and employ EGNN as the backbone of \method{}, named \method{}-EGNN. For this experiment, we report the F1 score and \% Error for all used models, where \% Error is defined as the percentage of incorrectly predicted edges relative to the total number of potential edges. EGNN is implemented based on the implementation details provided in \cite{satorras2021n}, which is used as the coupling function for message passing \cite{rusch2022graph}. Then, we use the ODE solvers provided in the torchdiffeq library for solving Eq. \ref{ode1} and Eq. \ref{ode2}.

\subsubsection{Results.} Table \ref{tab:autoencoder1} depicts the overall results of all methods on the Community and Erdos \& Renyi datasets. It is evident that our \method{}-EGNN can outperform all other methods across these two datasets, supporting the general effectiveness of \method{}. It is also worth noting that \method{}-EGNN utilizes EGNN as its backbone; the second-order information in \method{}-EGNN helps EGNN diminish the \% Error and increase the F1 score on these two datasets. Additionally, the experimental results demonstrate that \method{}-EGNN remains E(n)-equivariant to the noise distribution, provides the best reconstruction of the adjacency matrix ${A}_{ij}$ on a large-scale graph dataset, and enhances performance when embedding graphs into a continuous space as a set of nodes in dimension n. 

Then, we investigate the performance of \method{}-EGNN when the number of layers increases (up to 16) compared to EGNN, as shown in Table \ref{tab:autoencoder2}. We observe that a higher layer can lead to a lower F1 score and higher \% Error for the baseline EGNN model. Additionally, when the number of layers reaches eight on the Community dataset and 12 on the Erdos \& Renyi dataset, gradient explosion occurs, causing EGNN to produce infinite values. This phenomenon demonstrates that the gradient explosion occurs as the number of EGNN layers increases. On the contrary, our proposed \method{}-EGNN remains stable on both the Erdos \& Renyi and Community datasets, even with up to 16 layers, supporting our hypothesis that gradient explosion is mitigated when EGNN is implemented within \method{}.

\subsection{Molecular Data}
Finally, we evaluate \method{} against state-of-the-art baseline methods in molecular data, aiming to predict molecular chemical properties on 3D molecular graphs.

\subsubsection{Dataset.} The QM9 dataset is widely used in computational chemistry, which is specifically designed for the development and testing of quantum chemistry methods. It consists of approximately 134,000 stable small organic molecules made up of C, H, O, N, and F. Each molecule in the dataset contains up to nine non-hydrogen atoms. The dataset includes a diverse set of molecular properties that have been pre-computed using high-level quantum mechanical theories. We leveraged the QM9 dataset partitions from \cite{anderson2019cormorant} and sampled 100,000 molecules for training, 18,000 for validation, and 13,000 for testing \cite{brandstetter2021geometric}. Additionally, all properties have been normalized by subtracting the mean and dividing by the Mean Absolute Deviation \cite{satorras2021n}.

\subsubsection{Baselines.} Various widely used methods for molecule property prediction are deployed for comparison, such as DimeNet++  \cite{gasteiger_dimenetpp_2020}, SphereNet  \cite{liu2021spherical}, PaiNN  \cite{schutt2021equivariant}, PamNet  \cite{zhang2023universal}, and LEFTNet  \cite{du2024new}. Additionally, we deploy GMN ~\cite{huang2021equivariant} and SEGNO ~\cite{liu2024segno} on the molecule property prediction task.

\begin{table}[t]
\centering
\tabcolsep=1.8pt
\setlength{\abovecaptionskip}{0.2cm}
\caption{Performance comparison on the QM9 dataset. Numbers are reported for Mean Absolute Error (MAE) between model predictions and ground truth. ‘-' denotes no results are reported in the referred papers for the corresponding properties. The best performance is bold.}
\label{tab:table_5}
\resizebox{0.9\textwidth}{!}{
\begin{tabular}{lcccccccccccc}
\toprule
\multicolumn{1}{l}{Task} & \multicolumn{1}{c}{$\alpha$} & \multicolumn{1}{c}{$\Delta\epsilon$} & \multicolumn{1}{c}{$\epsilon_{\text{HOMO}}$} & \multicolumn{1}{c}{$\epsilon_{\text{LUMO}}$} & \multicolumn{1}{c}{$\mu$} & \multicolumn{1}{c}{$C_v$} & \multicolumn{1}{c}{$G$} & \multicolumn{1}{c}{$H$} & \multicolumn{1}{c}{$R^2$} & \multicolumn{1}{c}{$U$} & \multicolumn{1}{c}{$U_0$} & \multicolumn{1}{c}{ZPVE} \\ 
\multicolumn{1}{l}{Units} & bohr$^3$ & meV & meV & meV & D & cal/mol K & meV & meV & bohr$^3$ & meV & meV & meV \\ 
\midrule
NMP & .092 & 69 & 43 & 38 & .030 & .040 & 19 & 17 & .180 & 20 & 20 & 1.50 \\
SchNet & .235 & 63 & 41 & 34 & .033 & .033 & 14 & 14 & .073 & 19 & 14 & 1.70 \\
Cormorant & .085 & 61 & 34 & 38 & .038 & .026 & 20 & 21 & .961 & 21 & 22 & 2.02 \\
L1Net & .088 & 68 & 46 & 35 & .043 & .031 & 14 & 14 & .354 & 21 & 13 & 1.56 \\
LieConv & .084 & 49 & 30 & 25 & .032 & .038 & 22 & 24 & .800 & 19 & 19 & 2.28 \\
TFN & .223 & 58 & 40 & 38 & .064 & .101 & - & - & - & - & - & - \\
SE(3)-Tr. & .142 & 53 & 35 & 33 & .051 & .054 & - & - & - & - & - & - \\
\bottomrule
DimeNet++ & .047 & 45 & 27 & 24 & .028 & .027 & 11 & 11 & .573 & 11 & 11 & 1.49 \\
PaiNN & .049 & 49 & 29 & 23 & .029 & .028 & 11 & 10 & \textbf{.088} & 10 & 10 & 1.46 \\
SphereNet & .056 & 54 & 33 & 24 & .031 & .027 & {10} & {10} & .472 & 10 & {9} & \textbf{1.44} \\
PAMNet & .059 & 48 & 28 & 23 & .027 & .025 & 12 & 12 & .990 & 12 & 12 & 1.47 \\
LEFTNet & \textbf{.046} & 44 & 27 & 22 & \textbf{.022} & \textbf{.020} & 14 & 14 & .169 & 15 & 14 & 1.61 \\
EGNN & .071 & 48 & 29 & 25 & .028 & .032 & 12 & 12 & .103 & 12 & 12 & 1.60 \\
SEGNN & .077 & 46 & 28 & 24 & .033 & .035 & 12 & 11 & .899 & 11 & 12 & 1.65 \\
\bottomrule
SEGNO & .243 & 86 & 60 & 52 & .173 & .056 & 25 & 26 & .261 & 30 & 27 & 2.76 \\
GMN & .085 & 49 & 30 & 24 & .037 & .033 & 24 & 23 & .110 & 13 & 13 & 1.59 \\
\method{}-EGNN & .067 & \textbf{43} & \textbf{26} & \textbf{22} & .026 & .031 & 11 & 11 & .092 & 11 & 11 & 1.54 \\
\method{}-SEGNN & .072 & 50 & 28 & 26 & .031 & .033 & \textbf{10} & \textbf{10} & .880 & \textbf{10} & \textbf{9} & 1.62 \\
\bottomrule
\end{tabular}
}
\end{table}

\subsubsection{Experimental Setting.} Due to that DimeNet++  \cite{gasteiger_dimenetpp_2020}, SphereNet  \cite{liu2021spherical}, PaiNN  \cite{schutt2021equivariant}, PamNet  \cite{zhang2023universal}, and LEFTNet  \cite{du2024new} are trained under different train/val/test partitions, we rerun these six baselines, and report the results for a fair comparison. We rerun EGNN \cite{satorras2021n} and SEGNN \cite{brandstetter2021geometric} under the same experimental settings. All these baselines are evaluated on NVIDIA A100 GPUs. The experimental settings are as follows: Adam optimizer with learning rate ${1 \times 10^{-3}}$, the number of epochs 1,000, input embedding dim 128, hidden embedding dim 128, output embedding dim 128, batch size 128, weight decay ${1 \times 10^{-12}}$, and layer number 7. 

\subsubsection{Implementation details.} We build upon the experimental setting introduced in \cite{brandstetter2021geometric,satorras2021n}, where we aim to predict a variety of chemical properties for each of the molecules through regression. We optimized and reported the Mean Absolute Error (MAE) between predictions and ground truth. In this section, we employ EGNN and SEGNN as the backbone of the \method{} method, named \method{}-EGNN and \method{}-SEGNN, respectively. In particular, EGNN and SEGNN are implemented based on the implementation details provided in \cite{satorras2021n, brandstetter2021geometric}, which are used as the coupling function for message passing \cite{rusch2022graph}. Again, we use the ODE solvers provided in the torchdiffeq library for solving Eq. \ref{ode1} and Eq. \ref{ode2}.

\subsubsection {Results.} We use Table \ref{tab:table_5} to show the overall experimental results of all methods on the QM9 dataset. Compared to the baseline methods, our \method{}-EGNN achieves the best results for three energy variables ($\Delta\epsilon$, $\epsilon_{\text{HOMO}}$, and $\epsilon_{\text{LUMO}}$), while \method{}-SEGNN achieves the best results for four energy variables ($G$, $H$, $U$, and $U_0$). Additionally, we observe that incorporating second-order information into EGNN and SEGNN enables \method{}-EGNN and \method{}-SEGNN to outperform their original versions in most cases. In addition, \method{}-EGNN and \method{}-SEGNN outperform GMN and SEGNO in all tasks, which highlights the effectiveness of our approach. Therefore, we can conclude that \method{}-EGNN and \method{}-SEGNN can generate competitive results in most property prediction tasks while maintaining relatively simple model structures. To further clarify the conceptual distinction between our \method{} and specialized 3D architectures, we note that models such as PAMNet, SphereNet, PaiNN, and LEFTNet typically enhance expressiveness by explicitly incorporating higher-order geometric features or advanced attention mechanisms into their message-passing frameworks.
In contrast, our ODE-based approach adopts a fundamentally different philosophy: \method{} models the evolution of node features and coordinates as a continuous-time dynamic process governed by second-order ODEs, providing a principled, physics-inspired mechanism for capturing complex geometric dependencies without requiring manual engineering of geometric priors. Notably, these two paradigms are not mutually exclusive. Domain-specific geometric features or attention modules can be integrated into the ODE-based framework to enhance its expressiveness further. However, as demonstrated by the results in Table~\ref{tab:table_5}, our approach achieves highly competitive performance across a broad range of property prediction tasks even without explicit geometric augmentation. This underscores the generality and simplicity of ODE-based dynamics, as well as their potential to complement specialized 3D architectures.

\section{Conclusion}

In this paper, we introduced \method{}, a dual second-order equivariant graph neural network designed to address both feature and coordinate representations simultaneously through a graph ODE framework. Our approach not only maintains equivariant properties but also addresses two critical issues of deep graph neural networks: over-smoothing and gradient explosion problems. Empirical results on dynamic system modeling, graph auto-encoding, and molecular data tasks demonstrate the superior performance of our proposed \method{} over existing methods, especially in terms of data efficiency and stability across different layer numbers of GNN. These results underscore \method{}’s capacity to model complex physical and molecular systems effectively. Future work could explore extending \method{}’s application to other domains that benefit from high-order motion laws or developing variations that address specific challenges in large-scale or sparse datasets. Additionally, further integration with higher-order dynamics could reveal more insights into complex graph-structured data, enhancing its versatility across a broader range of tasks.

\bibliographystyle{plain}
\bibliography{references}

\begin{thebibliography}{10}

\bibitem{ai2023gcn}
Wei Ai, Yuntao Shou, Tao Meng, Nan Yin, and Keqin Li.
\newblock Der-gcn: Dialogue and event relation-aware graph convolutional neural network for multimodal dialogue emotion recognition.
\newblock {\em arXiv preprint arXiv:2312.10579}, 2023.

\bibitem{anderson2019cormorant}
Brandon Anderson, Truong~Son Hy, and Risi Kondor.
\newblock Cormorant: Covariant molecular neural networks.
\newblock In {\em Proceedings of the Conference on Neural Information Processing Systems}, volume~32, 2019.

\bibitem{battaglia2016interaction}
Peter Battaglia, Razvan Pascanu, Matthew Lai, Danilo Jimenez~Rezende, et~al.
\newblock Interaction networks for learning about objects, relations and physics.
\newblock In {\em Proceedings of the Conference on Neural Information Processing Systems}, volume~29, 2016.

\bibitem{brandstetter2021geometric}
Johannes Brandstetter, Rob Hesselink, Elise van~der Pol, Erik~J Bekkers, and Max Welling.
\newblock Geometric and physical quantities improve e (3) equivariant message passing.
\newblock In {\em Proceedings of the International Conference on Learning Representations}, 2021.

\bibitem{chen2018neural}
Ricky~TQ Chen, Yulia Rubanova, Jesse Bettencourt, and David~K Duvenaud.
\newblock Neural ordinary differential equations.
\newblock In {\em Proceedings of the Conference on Neural Information Processing Systems}, volume~31, 2018.

\bibitem{dasoulas2021lipschitz}
George Dasoulas, Kevin Scaman, and Aladin Virmaux.
\newblock Lipschitz normalization for self-attention layers with application to graph neural networks.
\newblock In {\em Proceedings of the International Conference on Machine Learning}, pages 2456--2466, 2021.

\bibitem{du2024new}
Yuanqi Du, Limei Wang, Dieqiao Feng, Guifeng Wang, Shuiwang Ji, Carla~P Gomes, Zhi-Ming Ma, et~al.
\newblock A new perspective on building efficient and expressive 3d equivariant graph neural networks.
\newblock {\em Advances in Neural Information Processing Systems}, 36, 2024.

\bibitem{erdds1959random}
P~ERDdS and A~R\&wi.
\newblock On random graphs i.
\newblock {\em Publ. math. debrecen}, 6(290-297):18, 1959.

\bibitem{fuchs2020se}
Fabian Fuchs, Daniel Worrall, Volker Fischer, and Max Welling.
\newblock Se (3)-transformers: 3d roto-translation equivariant attention networks.
\newblock In {\em Proceedings of the Conference on Neural Information Processing Systems}, pages 1970--1981, 2020.

\bibitem{gasteiger_dimenetpp_2020}
Johannes Gasteiger, Shankari Giri, Johannes~T. Margraf, and Stephan G{\"u}nnemann.
\newblock Fast and uncertainty-aware directional message passing for non-equilibrium molecules.
\newblock In {\em Proceedings of the International Conference on Neural Information Processing Systems Workshop}, 2020.

\bibitem{guo2022orthogonal}
Kai Guo, Kaixiong Zhou, Xia Hu, Yu~Li, Yi~Chang, and Xin Wang.
\newblock Orthogonal graph neural networks.
\newblock In {\em Proceedings of the AAAI Conference on Artificial Intelligence}, pages 3996--4004, 2022.

\bibitem{han2022equivariant}
Jiaqi Han, Wenbing Huang, Tingyang Xu, and Yu~Rong.
\newblock Equivariant graph hierarchy-based neural networks.
\newblock In {\em Proceedings of the Conference on Neural Information Processing Systems}, pages 9176--9187, 2022.

\bibitem{huang2021equivariant}
Wenbing Huang, Jiaqi Han, Yu~Rong, Tingyang Xu, Fuchun Sun, and Junzhou Huang.
\newblock Equivariant graph mechanics networks with constraints.
\newblock In {\em Proceedings of the International Conference on Learning Representations}, 2021.

\bibitem{huang2020learning}
Zijie Huang, Yizhou Sun, and Wei Wang.
\newblock Learning continuous system dynamics from irregularly-sampled partial observations.
\newblock In {\em Proceedings of the Conference on Neural Information Processing Systems}, pages 16177--16187, 2020.

\bibitem{huang2021coupled}
Zijie Huang, Yizhou Sun, and Wei Wang.
\newblock Coupled graph ode for learning interacting system dynamics.
\newblock In {\em Proceedings of the International ACM SIGKDD Conference on Knowledge Discovery \& Data Mining}, pages 705--715, 2021.

\bibitem{janson2011random}
Svante Janson, Tomasz Luczak, and Andrzej Rucinski.
\newblock {\em Random graphs}.
\newblock John Wiley \& Sons, 2011.

\bibitem{ju2024survey}
Wei Ju, Siyu Yi, Yifan Wang, Zhiping Xiao, Zhengyang Mao, Hourun Li, Yiyang Gu, Yifang Qin, Nan Yin, Senzhang Wang, et~al.
\newblock A survey of graph neural networks in real world: Imbalance, noise, privacy and ood challenges.
\newblock {\em arXiv preprint arXiv:2403.04468}, 2024.

\bibitem{kaplan2020scaling}
Jared Kaplan, Sam McCandlish, Tom Henighan, Tom~B Brown, Benjamin Chess, Rewon Child, Scott Gray, Alec Radford, Jeffrey Wu, and Dario Amodei.
\newblock Scaling laws for neural language models.
\newblock {\em arXiv preprint arXiv:2001.08361}, 2020.

\bibitem{keriven2022not}
Nicolas Keriven.
\newblock Not too little, not too much: a theoretical analysis of graph (over) smoothing.
\newblock In {\em Proceedings of the Conference on Neural Information Processing Systems}, pages 2268--2281, 2022.

\bibitem{kipf2018neural}
Thomas Kipf, Ethan Fetaya, Kuan-Chieh Wang, Max Welling, and Richard Zemel.
\newblock Neural relational inference for interacting systems.
\newblock In {\em Proceedings of the International Conference on Machine Learning}, pages 2688--2697, 2018.

\bibitem{kipf2016variational}
Thomas~N Kipf and Max Welling.
\newblock Variational graph auto-encoders.
\newblock {\em arXiv preprint arXiv:1611.07308}, 2016.

\bibitem{kipf2017semisupervised}
Thomas~N. Kipf and Max Welling.
\newblock Semi-supervised classification with graph convolutional networks.
\newblock In {\em Proceedings of the International Conference on Learning Representations}, 2017.

\bibitem{kohler2019equivariant}
Jonas K{\"o}hler, Leon Klein, and Frank No{\'e}.
\newblock Equivariant flows: sampling configurations for multi-body systems with symmetric energies.
\newblock {\em arXiv preprint arXiv:1910.00753}, 2019.

\bibitem{li2018deeper}
Qimai Li, Zhichao Han, and Xiao-Ming Wu.
\newblock Deeper insights into graph convolutional networks for semi-supervised learning.
\newblock In {\em Proceedings of the AAAI Conference on Artificial Intelligence}, 2018.

\bibitem{li2015gated}
Yujia Li, Daniel Tarlow, Marc Brockschmidt, and Richard Zemel.
\newblock Gated graph sequence neural networks.
\newblock {\em arXiv preprint arXiv:1511.05493}, 2015.

\bibitem{liu2024dynamic}
Ce~Liu, Jun Wang, Zhiqiang Cai, Yingxu Wang, Huizhen Kuang, Kaihui Cheng, Liwei Zhang, Qingkun Su, Yining Tang, Fenglei Cao, et~al.
\newblock Dynamic pdb: A new dataset and a se (3) model extension by integrating dynamic behaviors and physical properties in protein structures.
\newblock {\em arXiv preprint arXiv:2408.12413}, 2024.

\bibitem{liu2024segno}
Yang Liu, Jiashun Cheng, Haihong Zhao, Tingyang Xu, Peilin Zhao, Fugee Tsung, Jia Li, and Yu~Rong.
\newblock {SEGNO}: Generalizing equivariant graph neural networks with physical inductive biases.
\newblock In {\em Proceedings of the International Conference on Learning Representations}, 2024.

\bibitem{liu2021spherical}
Yi~Liu, Limei Wang, Meng Liu, Xuan Zhang, Bora Oztekin, and Shuiwang Ji.
\newblock Spherical message passing for 3d graph networks.
\newblock {\em arXiv preprint arXiv:2102.05013}, 2021.

\bibitem{luo2023hope}
Xiao Luo, Jingyang Yuan, Zijie Huang, Huiyu Jiang, Yifang Qin, Wei Ju, Ming Zhang, and Yizhou Sun.
\newblock Hope: High-order graph ode for modeling interacting dynamics.
\newblock In {\em Proceedings of the International Conference on Machine Learning}, pages 23124--23139, 2023.

\bibitem{meng2024deep}
Tao Meng, Yuntao Shou, Wei Ai, Nan Yin, and Keqin Li.
\newblock Deep imbalanced learning for multimodal emotion recognition in conversations.
\newblock {\em IEEE Transactions on Artificial Intelligence}, 2024.

\bibitem{min2020scattering}
Yimeng Min, Frederik Wenkel, and Guy Wolf.
\newblock Scattering gcn: Overcoming oversmoothness in graph convolutional networks.
\newblock In {\em Proceedings of the Conference on Neural Information Processing Systems}, pages 14498--14508, 2020.

\bibitem{norcliffe2020second}
Alexander Norcliffe, Cristian Bodnar, Ben Day, Nikola Simidjievski, and Pietro Li{\`o}.
\newblock On second order behaviour in augmented neural odes.
\newblock In {\em Proceedings of the Conference on Neural Information Processing Systems}, pages 5911--5921, 2020.

\bibitem{pang2023sa}
Jinhui Pang, Zixuan Wang, Jiliang Tang, Mingyan Xiao, and Nan Yin.
\newblock Sa-gda: Spectral augmentation for graph domain adaptation.
\newblock In {\em Proceedings of the 31st ACM International Conference on Multimedia}, pages 309--318, 2023.

\bibitem{poli2019graph}
Michael Poli, Stefano Massaroli, Junyoung Park, Atsushi Yamashita, Hajime Asama, and Jinkyoo Park.
\newblock Graph neural ordinary differential equations.
\newblock {\em arXiv preprint arXiv:1911.07532}, 2019.

\bibitem{rusch2022graph}
T~Konstantin Rusch, Ben Chamberlain, James Rowbottom, Siddhartha Mishra, and Michael Bronstein.
\newblock Graph-coupled oscillator networks.
\newblock In {\em Proceedings of the International Conference on Machine Learning}, pages 18888--18909, 2022.

\bibitem{sanchez2019hamiltonian}
Alvaro Sanchez-Gonzalez, Victor Bapst, Kyle Cranmer, and Peter Battaglia.
\newblock Hamiltonian graph networks with ode integrators.
\newblock {\em arXiv preprint arXiv:1909.12790}, 2019.

\bibitem{satorras2021n}
V{\i}ctor~Garcia Satorras, Emiel Hoogeboom, and Max Welling.
\newblock E (n) equivariant graph neural networks.
\newblock In {\em Proceedings of the International Conference on Machine Learning}, pages 9323--9332, 2021.

\bibitem{schutt2021equivariant}
Kristof Sch{\"u}tt, Oliver Unke, and Michael Gastegger.
\newblock Equivariant message passing for the prediction of tensorial properties and molecular spectra.
\newblock In {\em Proceedings of the International Conference on Machine Learning}, pages 9377--9388, 2021.

\bibitem{shou2023adversarial}
Yuntao Shou, Tao Meng, Wei Ai, Nan Yin, and Keqin Li.
\newblock Adversarial representation with intra-modal and inter-modal graph contrastive learning for multimodal emotion recognition.
\newblock {\em arXiv preprint arXiv:2312.16778}, 2023.

\bibitem{tang2024merging}
Anke Tang, Li~Shen, Yong Luo, Nan Yin, Lefei Zhang, and Dacheng Tao.
\newblock Merging multi-task models via weight-ensembling mixture of experts.
\newblock In {\em International Conference on Machine Learning}, 2024.

\bibitem{thomas2018tensor}
Nathaniel Thomas, Tess Smidt, Steven Kearnes, Lusann Yang, Li~Li, Kai Kohlhoff, and Patrick Riley.
\newblock Tensor field networks: Rotation-and translation-equivariant neural networks for 3d point clouds.
\newblock {\em arXiv preprint arXiv:1802.08219}, 2018.

\bibitem{ijcai2025p219}
Mengzhu Wang, Wenhao Ren, Yu~Zhang, Yanlong Fan, Dianxi Shi, Luoxi Jing, and Nan Yin.
\newblock Gaussian mixture model for graph domain adaptation.
\newblock In {\em Proceedings of the Thirty-Fourth International Joint Conference on Artificial Intelligence, {IJCAI-25}}, pages 1963--1972, 8 2025.

\bibitem{wang2025dynamically}
Yingxu Wang, Shiqi Fan, Mengzhu Wang, Siyang Gao, Siwei Liu, and Nan Yin.
\newblock Dynamically adaptive reasoning via llm-guided mcts for efficient and context-aware kgqa.
\newblock {\em arXiv preprint arXiv:2508.00719}, 2025.

\bibitem{wang2025nested}
Yingxu Wang, Mengzhu Wang, Zhichao Huang, Suyu Liu, and Nan Yin.
\newblock Nested graph pseudo-label refinement for noisy label domain adaptation learning.
\newblock {\em arXiv preprint arXiv:2508.00716}, 2025.

\bibitem{wanner1996solving}
Gerhard Wanner and Ernst Hairer.
\newblock {\em Solving ordinary differential equations II}, volume 375.
\newblock Springer Berlin Heidelberg New York, 1996.

\bibitem{xian2025molrag}
Ziting Xian, Jiawei Gu, Lingbo Li, and Shangsong Liang.
\newblock Molrag: unlocking the power of large language models for molecular property prediction.
\newblock In {\em Proceedings of the 63rd Annual Meeting of the Association for Computational Linguistics (Volume 1: Long Papers)}, pages 15513--15531, 2025.

\bibitem{xu2018powerful}
Keyulu Xu, Weihua Hu, Jure Leskovec, and Stefanie Jegelka.
\newblock How powerful are graph neural networks?
\newblock In {\em Proceedings of the International Conference on Learning Representations}, 2018.

\bibitem{yin2022dynamic}
Nan Yin, Fuli Feng, Zhigang Luo, Xiang Zhang, Wenjie Wang, Xiao Luo, Chong Chen, and Xian-Sheng Hua.
\newblock Dynamic hypergraph convolutional network.
\newblock In {\em 2022 IEEE 38th International Conference on Data Engineering (ICDE)}, pages 1621--1634. IEEE, 2022.

\bibitem{yin2022generic}
Nan Yin and Zhigang Luo.
\newblock Generic structure extraction with bi-level optimization for graph structure learning.
\newblock {\em Entropy}, 24(9):1228, 2022.

\bibitem{yinsport}
Nan Yin, Li~Shen, Chong Chen, Xian-Sheng Hua, and Xiao Luo.
\newblock Sport: A subgraph perspective on graph classification with label noise.
\newblock {\em ACM Transactions on Knowledge Discovery from Data}.

\bibitem{yin2022deal}
Nan Yin, Li~Shen, Baopu Li, Mengzhu Wang, Xiao Luo, Chong Chen, Zhigang Luo, and Xian-Sheng Hua.
\newblock Deal: An unsupervised domain adaptive framework for graph-level classification.
\newblock In {\em Proceedings of the 30th ACM International Conference on Multimedia}, pages 3470--3479, 2022.

\bibitem{yin2023coco}
Nan Yin, Li~Shen, Mengzhu Wang, Long Lan, Zeyu Ma, Chong Chen, Xian-Sheng Hua, and Xiao Luo.
\newblock Coco: A coupled contrastive framework for unsupervised domain adaptive graph classification.
\newblock In {\em International Conference on Machine Learning}, pages 40040--40053. PMLR, 2023.

\bibitem{yin2025dream}
Nan Yin, Li~Shen, Mengzhu Wang, Xinwang Liu, Chong Chen, and Xian-Sheng Hua.
\newblock Dream: a dual variational framework for unsupervised graph domain adaptation.
\newblock {\em IEEE Transactions on Pattern Analysis and Machine Intelligence}, 2025.

\bibitem{yin2023omg}
Nan Yin, Li~Shen, Mengzhu Wang, Xiao Luo, Zhigang Luo, and Dacheng Tao.
\newblock Omg: towards effective graph classification against label noise.
\newblock {\em IEEE Transactions on Knowledge and Data Engineering}, 2023.

\bibitem{yin2023messages}
Nan Yin, Li~Shen, Huan Xiong, Bin Gu, Chong Chen, Xian-Sheng Hua, Siwei Liu, and Xiao Luo.
\newblock Messages are never propagated alone: Collaborative hypergraph neural network for time-series forecasting.
\newblock {\em IEEE Transactions on Pattern Analysis and Machine Intelligence}, 2023.

\bibitem{ijcai2025p396}
Nan Yin, Xiao Teng, Zhiguang Cao, and Mengzhu Wang.
\newblock Coupling category alignment for graph domain adaptation.
\newblock In {\em Proceedings of the Thirty-Fourth International Joint Conference on Artificial Intelligence, {IJCAI-25}}, pages 3561--3569, 8 2025.

\bibitem{yin2024continuous}
Nan Yin, Mengzhu Wan, Li~Shen, Hitesh~Laxmichand Patel, Baopu Li, Bin Gu, and Huan Xiong.
\newblock Continuous spiking graph neural networks.
\newblock {\em arXiv preprint arXiv:2404.01897}, 2024.

\bibitem{yin2023dream}
Nan Yin, Mengzhu Wang, Zhenghan Chen, Li~Shen, Huan Xiong, Bin Gu, and Xiao Luo.
\newblock Dream: Dual structured exploration with mixup for open-set graph domain adaption.
\newblock In {\em The Twelfth International Conference on Learning Representations}, 2023.

\bibitem{you2018graphrnn}
Jiaxuan You, Rex Ying, Xiang Ren, William Hamilton, and Jure Leskovec.
\newblock Graphrnn: Generating realistic graphs with deep auto-regressive models.
\newblock In {\em Proceedings of the International Conference on Machine Learning}, pages 5708--5717, 2018.

\bibitem{zhang2022would}
Bowen Zhang, Daijun Ding, Liwen Jing, Genan Dai, and Nan Yin.
\newblock How would stance detection techniques evolve after the launch of chatgpt?
\newblock {\em arXiv preprint arXiv:2212.14548}, 2022.

\bibitem{zhang2023investigating}
Bowen Zhang, Xianghua Fu, Daijun Ding, Hu~Huang, Genan Dai, Nan Yin, Yangyang Li, and Liwen Jing.
\newblock Investigating chain-of-thought with chatgpt for stance detection on social media.
\newblock {\em arXiv preprint arXiv:2304.03087}, 2023.

\bibitem{zhang2023universal}
Shuo Zhang, Yang Liu, and Lei Xie.
\newblock A universal framework for accurate and efficient geometric deep learning of molecular systems.
\newblock {\em Scientific Reports}, 13(1):19171, 2023.

\end{thebibliography}

\end{document}